\PassOptionsToPackage{table}{xcolor}
\documentclass{article} 
\usepackage{tencent_ailab_tech_report}

\usepackage[T1]{fontenc} 
\usepackage[utf8]{inputenc} 

\usepackage{amsmath}
\usepackage{amssymb}
\usepackage{mathtools}
\usepackage{amsthm}

\usepackage[colorlinks,
            linkcolor=purple, 
            anchorcolor=blue,
            citecolor=blue!60!black
            ]{hyperref}
\usepackage{url}
\usepackage{enumitem}
\usepackage{booktabs} 
\usepackage{array}     
\usepackage{multirow}
\usepackage[utf8]{inputenc}
\usepackage{tabularx, makecell}
\usepackage[table]{xcolor}   
\usepackage{caption}
\usepackage{placeins} 
\usepackage{wrapfig} 
\usepackage{graphicx}   
\usepackage{listings}
\usepackage{hyperref}
\usepackage{soul}
\usepackage[most]{tcolorbox}

\usepackage{pifont}    
\definecolor{rowgray}{gray}{0.95}
\usepackage[T1]{fontenc}
\usepackage{amsthm} 
\usepackage{listings}
\usepackage{xcolor}             
\usepackage{threeparttable}

\definecolor{lightgray}{gray}{0.93}
\definecolor{lightblue}{RGB}{220,235,247}

\newtheorem{theorem}{Theorem}
\newtheorem{lemma}{Lemma}
\newtheorem{corollary}{Corollary}
\newtheorem{assumption}{Assumption}

\newcommand{\sert}[1]{\cellcolor{lightblue}#1}
\newcommand{\pos}[1]{\textcolor{red}{+{#1}}}
\newcommand{\negcolor}[1]{\textcolor{blue}{#1}}

\definecolor{mydarkblue}{rgb}{0,0.08,0.45}
\definecolor{mydarkgreen}{HTML}{32612D}
\definecolor{myblue}{HTML}{3b75c3}
\definecolor{myred}{HTML}{E33222}
\definecolor{mygreen}{HTML}{438773}
\definecolor{mymaroon}{RGB}{142,27,19}
\definecolor{maroon}{HTML}{800000}
\definecolor{mycite}{cmyk}{0.55,1,0,0.15}
\definecolor{codeblue}{rgb}{0.25,0.5,0.5}
\definecolor{codekw}{rgb}{0.85, 0.18, 0.50}
\definecolor{codegreen}{rgb}{0,0.6,0}
\definecolor{codegray}{rgb}{0.5,0.5,0.5}
\definecolor{codepurple}{rgb}{0.58,0,0.82}
\definecolor{backcolour}{rgb}{0.95,0.95,0.92}
\hypersetup{
    colorlinks=true,
    citecolor=codegreen,
          }

\newtcolorbox{bluebox}[1][]{
  enhanced,
  colframe=blue!40!gray,
  colback=white,
  coltitle=white,
  colbacktitle=blue!40!gray,
  width=\linewidth,
  arc=2mm,
  auto outer arc,
  boxrule=0.5pt,
  left=10pt,
  right=10pt,
  drop shadow={black!50!white},
  top=10pt,
  bottom=10pt,
  title={#1}, 
  fonttitle=\bfseries,
  title code={\node[rounded corners, fill=blue!75!black, draw=none, text=white] at (frame.title) {\textbf{#1}};}, 
  attach boxed title to top center={yshift=-2mm},
  boxed title style={sharp corners, size=small}
}

\usepackage[capitalize,noabbrev]{cleveref}

\theoremstyle{plain}
\theoremstyle{definition}
\theoremstyle{remark}

\usepackage[textsize=tiny]{todonotes}

\DeclareUnicodeCharacter{211D}{\ensuremath{\mathbb{R}}}
\DeclareUnicodeCharacter{2124}{\ensuremath{\mathbb{Z}}}
\DeclareUnicodeCharacter{21A6}{\ensuremath{\mapsto}}

\DeclareUnicodeCharacter{2264}{\ensuremath{\leq}}
\DeclareUnicodeCharacter{2265}{\ensuremath{\geq}}
\DeclareUnicodeCharacter{2192}{\ensuremath{\rightarrow}}
\DeclareUnicodeCharacter{2194}{\ensuremath{\leftrightarrow}}
\DeclareUnicodeCharacter{2227}{\ensuremath{\wedge}}
\DeclareUnicodeCharacter{22A2}{\ensuremath{\vdash}}
\DeclareUnicodeCharacter{2080}{\ensuremath{_0}}
\DeclareUnicodeCharacter{2081}{\ensuremath{_1}}
\DeclareUnicodeCharacter{2082}{\ensuremath{_2}}
\DeclareUnicodeCharacter{2083}{\ensuremath{_3}}
\DeclareUnicodeCharacter{2084}{\ensuremath{_4}}
\DeclareUnicodeCharacter{2085}{\ensuremath{_5}}
\DeclareUnicodeCharacter{2086}{\ensuremath{_6}}
\DeclareUnicodeCharacter{2087}{\ensuremath{_7}}
\DeclareUnicodeCharacter{2088}{\ensuremath{_8}}
\DeclareUnicodeCharacter{2089}{\ensuremath{_9}}
\DeclareUnicodeCharacter{2200}{\ensuremath{\forall}}
\DeclareUnicodeCharacter{2203}{\ensuremath{\exists}}
\DeclareUnicodeCharacter{2228}{\ensuremath{\lor}}
\DeclareUnicodeCharacter{2260}{\ensuremath{\neq}}
\DeclareUnicodeCharacter{230B}{\ensuremath{\rfloor}} 
\DeclareUnicodeCharacter{2223}{\ensuremath{\mid}}    
\DeclareUnicodeCharacter{230A}{\ensuremath{\lfloor}} 
\DeclareUnicodeCharacter{2090}{\ensuremath{_a}}  
\DeclareUnicodeCharacter{2091}{\ensuremath{_e}}  
\DeclareUnicodeCharacter{2092}{\ensuremath{_o}}  
\DeclareUnicodeCharacter{2093}{\ensuremath{_x}}  
\DeclareUnicodeCharacter{2211}{\ensuremath{\sum}}        
\DeclareUnicodeCharacter{2115}{\ensuremath{\mathbb{N}}}  

\DeclareUnicodeCharacter{2208}{\ensuremath{\in}} 
\title{Evolving Language Models without Labels:\\ \textit{Majority Drives Selection, Novelty Promotes Variation}}

\author{
Yujun Zhou$^{1,2}$$^\dagger$\thanks{Work done during Yujun's Internship at Tencent AI Lab.}\;, 
Zhenwen Liang$^1$ $^\dagger$, 
Haolin Liu$^{1,3}$, 
Wenhao Yu$^1$, 
Kishan Panaganti$^1$, \\
\textbf{
Linfeng Song$^1$,
Dian Yu$^1$,
Xiangliang Zhang$^2$, 
Haitao Mi$^1$,
Dong Yu$^1$}
\vspace{1em}\\
$^1$Tencent AI Lab, 
$^2$University of Notre Dame, 
$^3$University of Virginia\\
\vspace{0.1cm} \\ \vspace{0.1cm}
$^\dagger$ Core contributors \quad
\\
Correspondence to: \texttt{yzhou25@nd.edu}, \texttt{zhenwzliang@global.tencent.com}
}

\colmfinalcopy 
\begin{document}

\maketitle

\begin{abstract}
Large language models (LLMs) can improve through reinforcement learning with verifiable rewards (RLVR), but many settings lack reference answers or external verifiers. Label-free self-improvement instead constructs training signals from the model's own outputs. Because the same policy generates both the candidate responses and the signals derived from them, each update also changes the candidates available for future training, creating a self-reinforcing feedback loop. We identify a failure mode of this loop, which we call \emph{selection-induced diversity collapse}: repeated self-selection can progressively narrow the candidate pool and make alternative correct solutions harder to sample. To address this problem, we propose \textsc{Evol-RL}, which balances \emph{selection} and \emph{variation}. Majority voting provides a stable selection signal, while embedding-based novelty refines credit among responses with the same selection status; entropy regularization and asymmetric clipping further support this process by maintaining diverse rollouts and preserving rare, high-advantage updates. Across multiple training distributions, model scales, backbones, and matched label-free baselines, \textsc{Evol-RL} consistently improves pass@$n$ while generally maintaining or improving pass@1. For example, when trained without labels on AIME24, Qwen3-4B-Base improves over TTRL on held-out AIME25 from 4.6\% to 17.1\% in pass@1 and from 18.5\% to 42.0\% in pass@16. Training analyses further show that when the current majority is wrong, \textsc{Evol-RL} better preserves alternative correct solutions.

\begin{center}

    \raisebox{-0.25ex}{\includegraphics[height=1em]{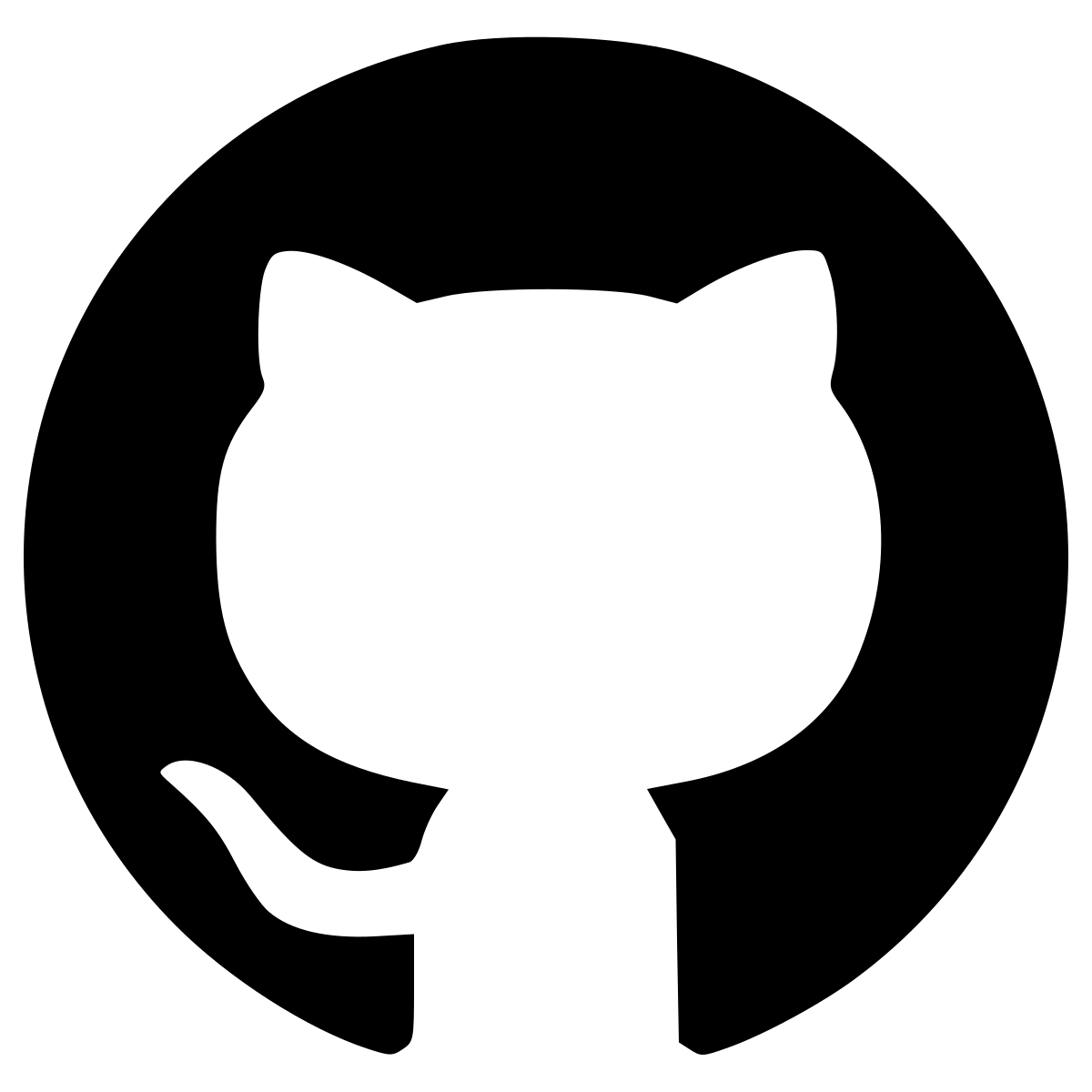}}\ \href{https://github.com/YujunZhou/EVOL-RL}{Code}~~~~~~
    \raisebox{-0.4ex}{\includegraphics[height=1em]{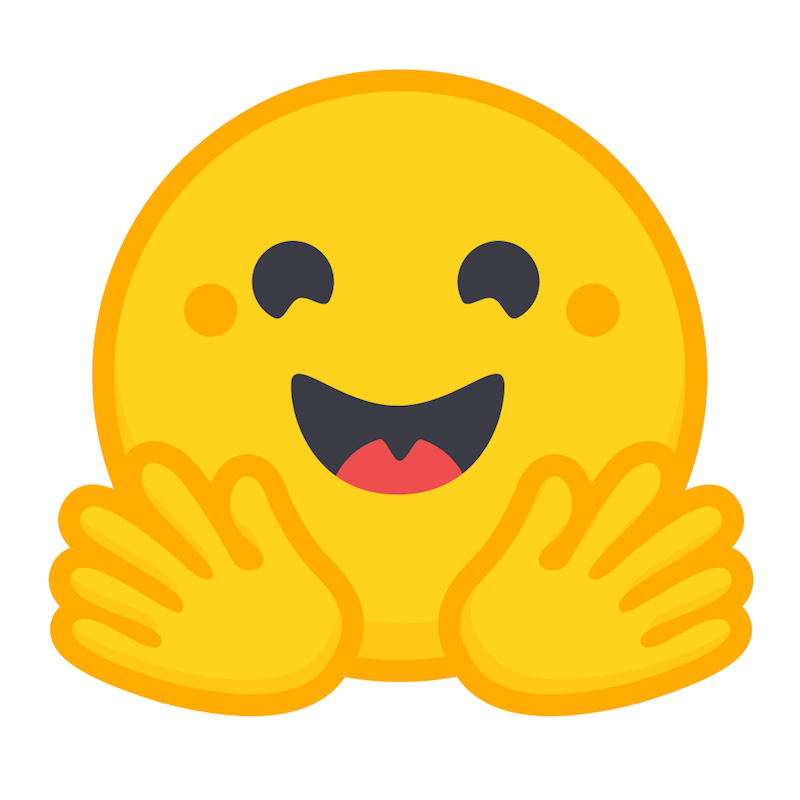}}\ \href{https://huggingface.co/collections/yujunzhou/evol-rl-68d8f3f7e2fadab49d6c3b9b}{Models}
\end{center}

\end{abstract}

\section{Introduction}

Reinforcement learning with verifiable rewards has advanced the reasoning capabilities of large language models (LLMs) \citep{jaech2024openai,guo2025deepseek,yang2025qwen3}, but continued improvement cannot always rely on reference answers or external judges. Label-free self-improvement instead derives supervision from the model's own outputs, using signals such as pseudo-labels, confidence, or consistency \citep{grandvalet2004semi,lee2013pseudo,zuo2025ttrl,shafayat2025can,li2025self}. Concretely, the current policy generates candidate responses, these responses are used to construct rewards or pseudo-labels, and the resulting signal updates the policy. The updated policy then generates the candidates used to construct supervision in the next round. As a result, each update changes not only what the model generates next, but also the candidate solutions from which its next training signal will be constructed.

This closed loop becomes problematic when the self-generated signal repeatedly favors responses that already agree with the model's dominant belief. Those responses become more likely under the updated policy, while alternative reasoning paths are sampled less often; future supervision is then constructed from an increasingly concentrated candidate set, further reinforcing the same modes. Related concerns about self-reinforcing feedback and learning from model-generated data have been studied in self-feedback, alignment, and recursive training \citep{liang2024internalconsistencyselffeedbacklarge,lee-etal-2024-aligning,ding2024sail,liang2025can,shumailov2024ai,li2026cyclical}. We refer to this failure mode as \emph{selection-induced diversity collapse}. Its central consequence is the \emph{loss of corrective alternatives}: alternative correct solutions can become progressively less likely to be sampled, making an erroneous consensus harder to overturn even as single-sample accuracy improves.

\begin{figure}[t]
    \centering
    \includegraphics[width=\textwidth]{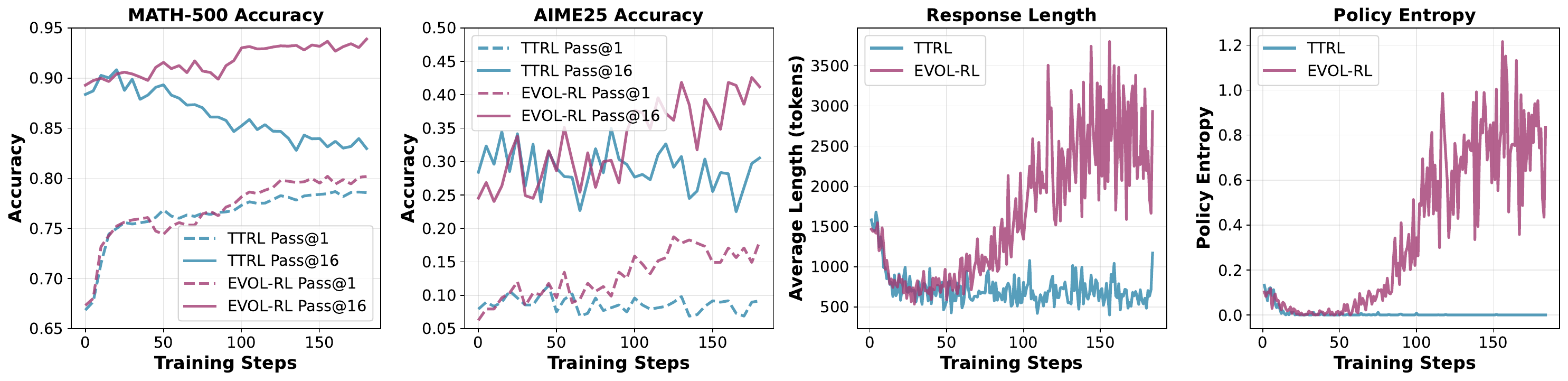}
    \caption{Training dynamics of TTRL and \textsc{Evol-RL} on Qwen3-4B-Base. TTRL improves pass@1, but pass@16, response length, and policy entropy decline; \textsc{Evol-RL} preserves broader coverage.}
    \label{fig:cog_collapse}
    \vspace{-1.1em}
\end{figure}

To make this failure mode concrete, we consider TTRL \citep{zuo2025ttrl}, a representative majority-vote approach to label-free self-improvement. TTRL uses the majority answer among the current policy's rollouts as a pseudo-label and reinforces responses that agree with it. It therefore directly instantiates the feedback loop described above: the policy determines the candidate set, the candidate set determines the supervision, and that supervision is then used to update the policy. As shown in Figure~\ref{fig:cog_collapse}, TTRL can improve single-sample accuracy while multi-sample coverage, response length, and policy entropy decline.

TTRL provides the clearest example, but the concern is broader: repeatedly reinforcing a model's belief can narrow the candidate set from which future supervision is constructed. In our matched comparisons, several other label-free objectives show similar pass@1--pass@$k$ trade-offs, suggesting that gains in pass@1 do not necessarily preserve alternative solutions for future correction.

The problem is not selection itself, but selection without a mechanism to preserve diversity. Label-free self-improvement needs a stable signal that favors promising solutions, while still keeping other plausible solutions available as the policy evolves. We describe these two roles as \emph{selection} and \emph{variation}. This balance mirrors the classical selection--variation principle in evolutionary computation \citep{holland1992adaptation,eiben2015introduction,lehman2011abandoning,pugh2016quality}. In our setting, variation is valuable not only for exploration, but also for preserving the candidate pool from which future supervision is constructed.

We instantiate this principle in \emph{EVolution-Oriented and Label-free Reinforcement Learning} (\textsc{Evol-RL}). Its core design separates \emph{selection} from \emph{variation}: majority voting first determines which responses are favored, and an embedding-based novelty signal then redistributes credit among responses with the same majority status. This preserves variation without allowing novelty to override the selection signal. Entropy regularization and asymmetric clipping support this process by maintaining a supply of alternative rollouts and preserving updates from rare, high-advantage samples \citep{haarnoja2018soft,yu2025dapo}. Together, these mechanisms keep alternative reasoning paths available while the model continues to learn from its own outputs.

\textbf{Contributions.}
\textbf{(1)} We diagnose a failure mode of label-free self-improvement in which self-generated selection progressively narrows exploration, making alternative correct solutions harder to recover and eventually leading to diversity collapse.
\textbf{(2)} We frame label-free self-improvement as a balance between selection and variation, instantiate this principle in \textsc{Evol-RL}, and provide an idealized analysis showing that variation promotes broader coverage of correct reasoning modes.
\textbf{(3)} Across multiple training settings, model scales, backbones, and matched baselines, \textsc{Evol-RL} reverses the pass@$n$ degradation observed in prior label-free methods, preserves alternative reasoning paths, and improves held-out reasoning performance.

\section{Related Works} 
\vspace{-1mm}
\textbf{Enhancing Reasoning in LLMs.} Significant progress in LLM reasoning has been driven by RLVR \citep{jaech2024openai, guo2025deepseek,yang2025qwen3,yu2025dapo,xiong2025rag,dai2025r1,wang2025causally,zhuang2025exploring}, which fine-tunes models using RL on tasks where an automated verifier can confirm the correctness of the final answer, such as mathematics and coding  \citep{zeng2025simplerl,wang2025beyond,wang2025adareasoner,cui2025entropy,huang2025r,dai2025cde,zheng2025parallel,zhou2025dissecting,zheng2025learning,fang2025serl,liu2025stable}. While highly effective, the reliance of RLVR on external verifiers restricts its applicability to domains with deterministic, easily checkable solutions \citep{zhao2025one,zhao2025majority,zhou2025reinforcing,zhou2024defending,liang2025clue}. Our work instead studies self-improvement when such external verification is unavailable.

\textbf{Label-Free Adaptation and Self-Improvement.}
To reduce reliance on external verifiers, label-free methods derive training
signals from the model's own outputs using confidence, entropy, self-consistency,
or agreement across multiple samples
\citep{prabhudesai2025maximizing,agarwal2025unreasonable,zhao2025learning,
zhang2025right,shafayat2025can,chung2025modifying,zhang2025consistent,
zuo2025ttrl}. TTRL, for example, uses the majority-voted answer as a pseudo-label
for RL updates \citep{zuo2025ttrl}. Because these signals are generated by the
current policy itself, repeated reinforcement can narrow the candidate pool and
make alternative correct solutions harder to sample. Related work improves
exploration through entropy regularization, asymmetric clipping, or other
exploration objectives \citep{cui2025entropy,park2025clip,liu2025ettrl}, while
some approaches introduce separately trained evaluators
\citep{li2025jointly,pang2024language}. \textsc{Evol-RL} instead explicitly
separates selection from variation: majority voting determines selection status,
while embedding-based novelty refines credit only among responses with the same
status.

\section{Method}
\begin{figure*}[t]
    \centering
    \includegraphics[width=\textwidth]{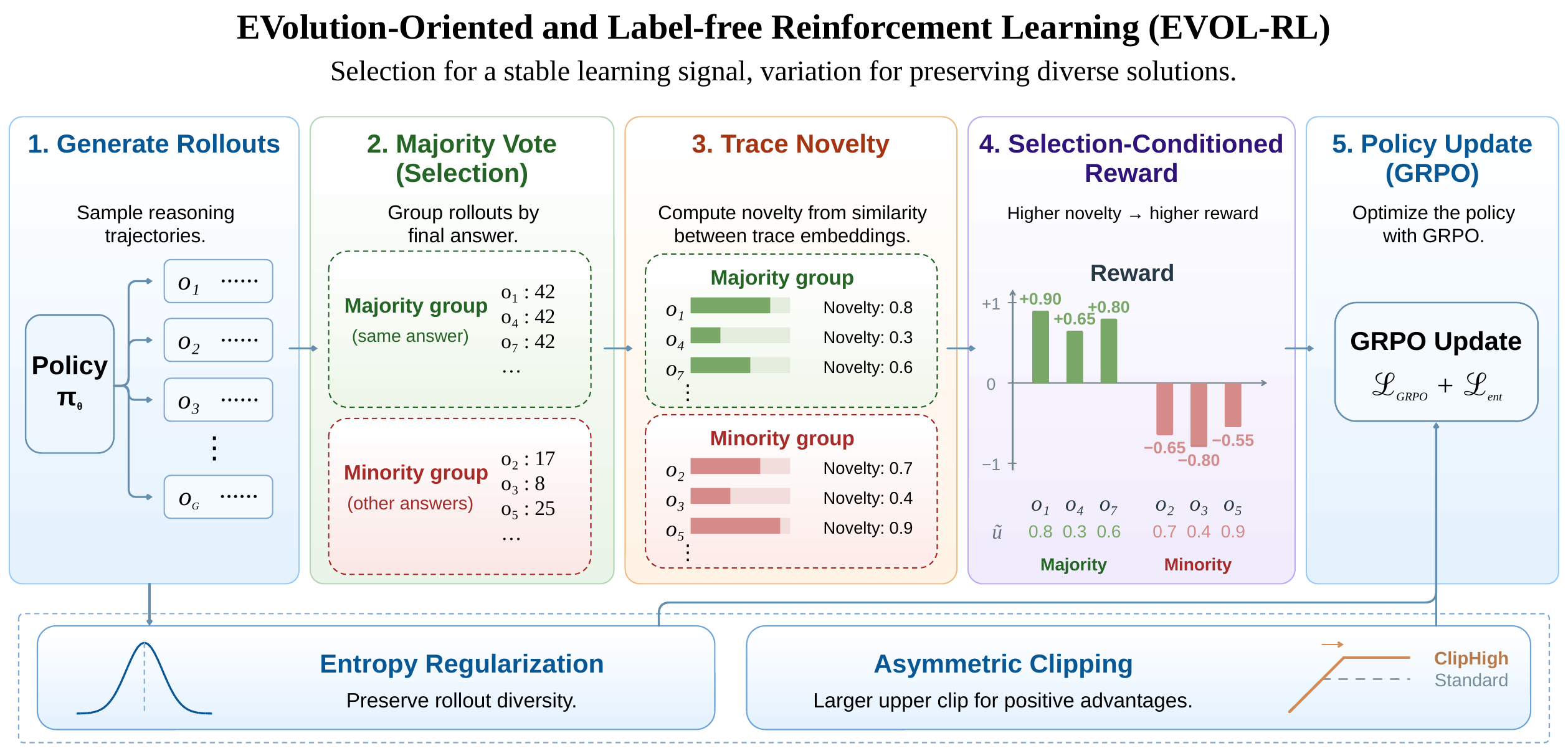}
    \vspace{-1.1em}
   \caption{Overview of \textsc{Evol-RL}. Majority voting provides the selection signal, while trace novelty refines rewards within each selection group. Entropy regularization maintains rollout diversity, and asymmetric clipping preserves rare, high-advantage updates. Values are illustrative.}
    \label{fig:method}
    \vspace{-1.1em}
\end{figure*}

Our approach is illustrated in Figure~\ref{fig:method}. \textsc{Evol-RL} uses
Group Relative Policy Optimization (GRPO) \citep{shao2024deepseekmath} for
policy optimization, while separating two roles in the training signal:
majority voting provides selection, and trace novelty preserves variation
within each selection group. Entropy regularization and asymmetric clipping
support this process by maintaining diverse rollouts and preserving
high-advantage updates.

\subsection{Optimization with GRPO}

GRPO is a policy-gradient algorithm designed for fine-tuning LLMs without a separate value function. Its central idea is to evaluate each sampled response relative to a group of its peers generated for the same prompt.
For a prompt $\mathbf{q}$, a policy LLM $\pi_{\theta_{\text{old}}}$ generates a group of $G$ responses $\{\mathbf o_1,\ldots,\mathbf o_G\}$. Each response $\mathbf o_i$ receives a scalar reward $r_i$. Rewards within the group are normalized with a z-score to obtain a response-level advantage:
\vspace{-1mm}
$$
\hat{A}_i=\frac{r_i-\text{mean}(r_1,\ldots,r_G)}{\text{std}(r_1,\ldots,r_G)},
$$
\vspace{-1mm}
The policy is optimized with a clipped surrogate objective:
{\small
\begin{equation}
\tfrac{1}{G}\sum_{i=1}^{G}\tfrac{1}{|o_i|}\sum_{t=1}^{|o_i|} \min\!\Big\{ \tfrac{\pi_{\theta}(o_{i,t}\mid \mathbf{q},\mathbf{o}_{i,<t})}{\pi_{\theta_{\mathrm{old}}}(o_{i,t}\mid \mathbf{q},\mathbf{o}_{i,<t})}\,\hat{A}_{i,t},\; \operatorname{clip}\!\Big(\tfrac{\pi_{\theta}(o_{i,t}\mid \mathbf{q},\mathbf{o}_{i,<t})}{\pi_{\theta_{\mathrm{old}}}(o_{i,t}\mid \mathbf{q},\mathbf{o}_{i,<t})},\,1-\epsilon_{\mathrm{low}},\,1+\epsilon_{\mathrm{high}}\Big)\,\hat{A}_{i,t} \Big\}
\end{equation}
}


\subsection{Reward Design: Selection-Conditioned Variation}

Our reward design assigns distinct roles to \emph{selection} and
\emph{variation}. Majority voting first determines whether each response agrees
with the current consensus. Novelty then refines credit \emph{within} the
majority and minority groups, rather than competing with the selection signal.
Among majority responses, more distinct reasoning traces receive higher rewards
than near-duplicates. Among minority responses, higher novelty leads to a less
negative reward, preserving alternative trajectories without rewarding
disagreement with the majority itself. Non-overlapping reward bands ensure that
selection always takes priority over variation.

\paragraph{Reward Formulation.}
For each prompt, the policy samples $G$ responses
$\{o_i\}_{i=1}^{G}$. Each response is scored using three components:

\textbf{1. Validity.}
The response must provide a numeric final answer in a
\verb|\boxed{·}| format. Responses that fail this check are treated as invalid.

\textbf{2. Majority (Selection).}
Among valid responses, we identify the majority-voted final answer and assign
a binary label $y_i\in\{+1,-1\}$ according to whether response $o_i$ agrees
with it. This label determines the response's selection group.

\textbf{3. Novelty (Variation).}
We embed the reasoning trace of each response and compute pairwise cosine
similarities. For response $o_i$, let $\bar{s}_i$ denote its mean similarity to
other responses in the same selection group, and let $m_i$ denote its maximum
similarity to any other response in the rollout set. We define
\[
u_i
=
1-\left(\alpha\bar{s}_i+(1-\alpha)m_i\right),
\qquad
\alpha\in[0,1],
\]
with $\alpha=0.5$ by default.

The two terms capture complementary forms of redundancy:
$\bar{s}_i$ measures how closely a response follows the common pattern within
its group, while $m_i$ detects near-duplicates. We then min--max normalize
$\{u_i\}$ separately within the majority and minority groups to obtain
$\tilde{u}_i\in[0,1]$. This keeps novelty relative to responses with the same
selection status.

\paragraph{Final Reward Mapping.}
We map selection status and normalized novelty into non-overlapping reward
bands:
\[
r_i =
\begin{cases}
-1, & \text{if invalid},\\[2pt]
0.5 + 0.5\,\tilde{u}_i \in [0.5,1],
    & \text{if } y_i=+1,\\[2pt]
-1 + 0.5\,\tilde{u}_i \in [-1,-0.5],
    & \text{if } y_i=-1.
\end{cases}
\]

Thus, any majority response receives a higher reward than any minority
response, while novelty determines the relative credit within each group.
This preserves the majority-based selection signal while discouraging
concentration on near-duplicate reasoning traces. Additional implementation
details are provided in Appendix~\ref{app:method_details}.

\subsection{Supporting Mechanisms}

The reward design determines how selection and variation allocate credit.
We use two supporting mechanisms to help preserve the resulting learning
signals during optimization.

First, we use an asymmetric clipping range
($\epsilon_{\mathrm{high}}>\epsilon_{\mathrm{low}}$)
\citep{yu2025dapo}. A larger upper clipping threshold allows
positive-advantage responses to increase in probability over a wider range
before being clipped, preserving stronger updates from rare high-advantage
samples.

Second, we add a token-level entropy regularizer to discourage premature
policy concentration and maintain a diverse supply of rollouts for subsequent
training steps. In loss notation, $\mathcal{L}_{\mathrm{GRPO}}$ denotes the
negative clipped surrogate, and we minimize the entropy loss
\[
\mathcal{L}_{\mathrm{ent}}(\theta)
=
-\lambda_{\mathrm{ent}}\,
\mathbb{E}_{o\sim\pi_\theta}
\left[
\frac{1}{|o|}
\sum_{t=1}^{|o|}
H\!\left(\pi_\theta(\cdot\mid o_{<t},x)\right)
\right],
\qquad
H(p)=-\sum_v p(v)\log p(v).
\]

Together, asymmetric clipping preserves strong positive updates, while entropy
regularization helps maintain the candidate diversity on which the
selection--variation reward operates.

\subsection{Theoretical Analysis}

To isolate the role of variation from pseudo-label noise and optimization
details, we analyze a simplified objective in which correctness is known and
variation is represented by pairwise trajectory dissimilarity:
\[
J_0(\pi)=\mathbb{E}_{y\sim\pi}[r(y)], \qquad
J_\lambda(\pi)
=
J_0(\pi)+\lambda\,\mathbb{E}_{y,y'\sim\pi}[d(y,y')],
\]
where $\lambda>0$ controls the strength of variation.

The key observation is that $J_0$ is indifferent once all probability lies on
correct solutions: collapsing onto one correct mode and covering many can
achieve the same objective value. A small positive $\lambda$ can break this tie
without overpowering correctness.

\begin{theorem}[Variation promotes mode coverage]
Suppose correct trajectories form $M\ge2$ reasoning modes with greater
cross-mode than within-mode dissimilarity. Let
$D_{\max}=\max_{y,y'}d(y,y')$. If
\[
0<\lambda<\frac{1}{2D_{\max}},
\]
then every global maximizer of $J_\lambda$ still places all probability on
correct trajectories, but assigns positive probability to every correct mode:
\[
w_m(\pi)=\sum_{y\in G_m}\pi(y)>0,\qquad m=1,\ldots,M.
\]
In contrast, $J_0$ admits optima concentrated on a single correct mode. Under
symmetric modes, $J_\lambda$ further selects the uniform distribution over all
correct trajectories.
\end{theorem}

Thus, the simplified analysis captures the principle behind our reward design:
once selection favors good solutions, variation breaks ties among them in favor
of broader mode coverage. Full assumptions and proofs are in
Appendix~\ref{app:theory}.

\section{Experiments}
\subsection{Experimental Setup}

\textbf{Training and evaluation.}
We train on the standard \textbf{MATH-TRAIN} set
\citep{hendrycks2021measuring} and, following TTRL
\citep{zuo2025ttrl}, on the smaller \textbf{MATH-500} and
\textbf{AIME24} sets \citep{li2024numinamath}. All training is fully
label-free: only problem statements are used, without reference answers or
solutions. We evaluate on \textbf{MATH-500}, \textbf{AIME24},
\textbf{AIME25}, \textbf{AMC} \citep{li2024numinamath}, and
\textbf{GPQA-Diamond (GPQA)} \citep{rein2024gpqa}. When evaluation uses the
same prompts seen during training, we treat the result as \emph{transductive
adaptation}; performance on unseen benchmarks measures held-out transfer.
Our main experiments compare \textsc{Evol-RL} with TTRL across
training distributions and model scales, while broader label-free methods are
compared under matched representative protocols in Section~\ref{sec:baseline_comparison}.
Full training details are provided in Appendix~\ref{app:implementation_details};
additional architecture, efficiency, similarity-validation, and sensitivity
results are reported in Appendix \ref{app:additional_experimental_results}.

\begin{table*}[t]
\centering
\renewcommand{\arraystretch}{1.25}
\setlength{\tabcolsep}{6pt}
\caption{TTRL and \textsc{Evol-RL} across training settings. Each cell reports pass@1/pass@16 averaged over 32 rollouts; $\Delta$ denotes \textsc{Evol-RL} minus TTRL.}
\vspace{-2mm}
\resizebox{0.88\textwidth}{!}{
\begin{tabular}{ccccccc}
\toprule
\textbf{Training Dataset} & \textbf{Model} & \textbf{MATH} & \textbf{AIME24} & \textbf{AIME25} & \textbf{AMC} & \textbf{GPQA} \\
\rowcolor{lightgray}
\midrule
\multicolumn{7}{c}{\textbf{Qwen3-4B-Base}} \\
\midrule
-- & Base Model & 67.4/89.6 & 10.0/32.4 & 5.5/30.0 & 39.3/75.2 & 34.4/85.6 \\
\midrule
\multirow{3}{*}{MATH-TRAIN}
& w/TTRL         & 75.4/86.9 & 12.1/23.2 & 6.8/28.6  & 42.5/75.2 & 36.5/81.4 \\
& \sert{w/\textsc{Evol-RL}}& \sert{80.0/93.3} & \sert{20.7/47.6} & \sert{17.5/39.9} & \sert{51.4/80.3} & \sert{37.2/88.7} \\
& $\Delta$       & \pos{4.6}/\pos{6.4} & \pos{8.6}/\pos{24.4} & \pos{10.7}/\pos{11.3} & \pos{8.9}/\pos{5.1} & \pos{0.7}/\pos{7.3} \\
\midrule
\multirow{3}{*}{MATH-500}
& w/TTRL         & 79.3/83.2 & 10.0/28.0 & 7.2/29.9  & 47.6/72.0 & 36.2/75.9 \\
& \sert{w/\textsc{Evol-RL}}& \sert{79.8/93.8} & \sert{19.0/43.2} & \sert{16.1/41.9} & \sert{50.3/82.2} & \sert{38.8/89.1} \\
& $\Delta$       & \pos{0.5}/\pos{10.6} & \pos{9.0}/\pos{15.2} & \pos{8.9}/\pos{12.0} & \pos{2.7}/\pos{10.2} & \pos{2.6}/\pos{13.2} \\
\midrule
\multirow{3}{*}{AIME24}
& w/TTRL         & 73.8/84.5 & 16.7/16.7 & 4.6/18.5  & 43.6/65.8 & 35.1/73.5 \\
& \sert{w/\textsc{Evol-RL}}& \sert{79.6/93.6} & \sert{20.6/40.9} & \sert{17.1/42.0} & \sert{49.9/80.9} & \sert{38.0/87.8} \\
& $\Delta$       & \pos{5.8}/\pos{9.1} & \pos{3.9}/\pos{24.2} & \pos{12.5}/\pos{23.5} & \pos{6.3}/\pos{15.1} & \pos{2.9}/\pos{14.3} \\
\rowcolor{lightgray}
\midrule
\multicolumn{7}{c}{\textbf{Qwen3-8B-Base}} \\
\midrule
-- & Base Model & 63.6/91.5 & 12.0/39.4 & 8.2/30.8 & 38.7/77.6 & 34.9/88.0 \\
\midrule
\multirow{3}{*}{MATH-TRAIN}
& w/TTRL         & 81.1/91.1 & 16.7/37.6 & 15.6/35.9 & 53.6/74.0 & 38.1/77.1 \\
& \sert{w/\textsc{Evol-RL}}& \sert{83.6/94.1} & \sert{26.0/51.7} & \sert{21.6/43.1} & \sert{55.5/86.1} & \sert{43.5/88.1} \\
& $\Delta$       & \pos{2.5}/\pos{3.0} & \pos{9.3}/\pos{14.1} & \pos{6.0}/\pos{7.2} & \pos{1.9}/\pos{12.1} & \pos{5.4}/\pos{11.0} \\
\midrule
\multirow{3}{*}{MATH-500}
& w/TTRL         & 85.7/91.9 & 17.7/40.1 & 16.5/34.3 & 51.1/79.1 & 43.5/84.0 \\
& \sert{w/\textsc{Evol-RL}}& \sert{84.7/95.1} & \sert{24.1/49.5} & \sert{20.2/44.4} & \sert{58.8/86.0} & \sert{43.9/92.2} \\
& $\Delta$       & \negcolor{-1.0}/\pos{3.2} & \pos{6.4}/\pos{9.4} & \pos{3.7}/\pos{10.1} & \pos{7.7}/\pos{6.9} & \pos{0.4}/\pos{8.2} \\
\midrule
\multirow{3}{*}{AIME24}
& w/TTRL         & 76.8/86.2 & 20.0/20.0 & 11.4/25.4 & 49.5/69.1 & 38.3/74.7 \\
& \sert{w/\textsc{Evol-RL}}& \sert{83.1/94.2} & \sert{25.4/38.1} & \sert{16.5/34.7} & \sert{54.4/85.8} & \sert{45.2/90.0} \\
& $\Delta$       & \pos{6.3}/\pos{8.0} & \pos{5.4}/\pos{18.1} & \pos{5.1}/\pos{9.3} & \pos{4.9}/\pos{16.7} & \pos{6.9}/\pos{15.3} \\
\bottomrule
\end{tabular}
}
\vspace{-0.8em}
\label{tab:main}
\end{table*}

\subsection{Main Results}

Table~\ref{tab:main} compares \textsc{Evol-RL} with TTRL across two model
scales and three label-free training sets. Averaged over all model--training--
evaluation combinations, \textsc{Evol-RL} improves pass@1 by 5.3 points and
pass@16 by 11.8 points over TTRL. The gains are particularly large on harder
benchmarks: Qwen3-4B trained on MATH-TRAIN gains 24.4 points in AIME24
pass@16, while training on AIME24 improves held-out AIME25 pass@16 by
23.5 points. The substantially larger gain in pass@16 is consistent with the
intended effect of preserving a broader set of solutions without sacrificing
their overall accuracy.

The improvements remain consistent across Qwen3-4B/8B and across
MATH-TRAIN, MATH-500, and AIME24, indicating that the benefit is not tied to a
particular model scale or training distribution. For example, Qwen3-4B trained
on MATH-500 improves AIME24 pass@16 from 28.0 to 43.2 and AIME25 from
29.9 to 41.9, with similar trends for the 8B model.

The same pattern holds on benchmarks that are always unseen during training.
Across the six training configurations, the average gains on AIME25 are
+7.8 points in pass@1 and +12.2 in pass@16; on GPQA, they are +3.2 and
+11.6 points, respectively. These results indicate that balancing selection
with variation improves not only adaptation to the training distribution but
also transfer to held-out reasoning tasks.

\subsection{Comparison with Other Label-Free Methods}
\label{sec:baseline_comparison}

\begin{wraptable}{r}{0.45\linewidth}
    \centering
    \vspace{-1.25em}
    \renewcommand{\arraystretch}{1.15}
    \setlength{\tabcolsep}{3.5pt}
    \caption{Matched comparison on Qwen3-4B-Base trained on MATH-500.
    Each cell reports pass@1/pass@16.}
    \vspace{-1mm}
    \resizebox{\linewidth}{!}{
    \begin{tabular}{lccc}
        \toprule
        \textbf{Method} & \textbf{MATH} & \textbf{AIME24} & \textbf{AIME25} \\
        \midrule
        Base Model
            & 67.4/89.6 & 10.0/32.4 & 5.5/30.0 \\
        TTRL
            & 79.3/83.2 & 10.0/28.0 & 7.2/29.9 \\
        EMPO
            & \textbf{79.9}/88.9 & 11.8/31.6 & 9.5/26.3 \\
        Co-Reward
            & 77.6/89.2 & 13.6/25.9 & 10.5/30.3 \\
        ECHO
            & 78.5/86.4 & 10.6/30.6 & 10.5/31.7 \\
        \sert{\textsc{Evol-RL}}
            & \sert{79.8/\textbf{93.8}}
            & \sert{\textbf{19.0/43.2}}
            & \sert{\textbf{16.1/41.9}} \\
        \bottomrule
    \end{tabular}}
    \label{tab:baseline_comparison}
\end{wraptable}

To assess whether the gains extend beyond TTRL, we compare \textsc{Evol-RL}
with several label-free self-improvement methods under a matched MATH-500
protocol using Qwen3-4B-Base: TTRL \citep{zuo2025ttrl}, EMPO
\citep{zhang2025right}, Co-Reward \citep{zhang2025co}, and the recent ECHO
\citep{zhao2026echo}. All methods use the same training data, rollout budget,
optimization settings, and evaluation protocol, so we treat these results as
matched-protocol comparisons rather than exact reproductions of independently
tuned pipelines.

As shown in Table~\ref{tab:baseline_comparison}, this tension is not unique to
TTRL. Several label-free baselines improve pass@1 while failing to preserve
pass@16, particularly on AIME24 and AIME25. In contrast, \textsc{Evol-RL}
improves both metrics across all three benchmarks, suggesting that explicitly
preserving variation helps avoid the loss of solution coverage during
self-improvement. Additional matched comparisons with EM-RL-Token/Sequence
\citep{agarwal2025unreasonable} and Self-Consistency
\citep{wang2022self,huang2023large} are reported in
Appendix~\ref{sec:comparison_other_baselines}.

\subsection{Ablation Study}
\begin{table*}[t]
\centering
\renewcommand{\arraystretch}{1.22}
\setlength{\tabcolsep}{6pt}
\caption{Performance of Qwen3-4B-Base with \textsc{Evol-RL} and its ablations. Each cell reports pass@1/pass@16 accuracy.}
\vspace{-2mm}
\label{tab:ablation}
\resizebox{0.88\textwidth}{!}{
\begin{tabular}{clccccc}
\toprule
\textbf{Training Dataset} & \textbf{Model} & \textbf{MATH} & \textbf{AIME24} & \textbf{AIME25} & \textbf{AMC} & \textbf{GPQA} \\
\midrule
\multirow{6}{*}{MATH-500}
& \sert{\textbf{w/\textsc{Evol-RL}}}            & \sert{\textbf{79.8}/\textbf{93.8}} & \sert{\textbf{19.0}/\textbf{43.2}} & \sert{\textbf{16.1}/\textbf{41.9}} & \sert{\textbf{50.3}/\textbf{82.2}} & \sert{\textbf{38.8}/\textbf{89.1}} \\
& \hspace{0.5em}-ClipHigh          & 75.1/91.8 & 12.2/31.8 & 11.4/31.3 & 42.7/73.9 & 32.3/81.8 \\
& \hspace{0.5em}-Ent               & 79.5/93.4 & 18.3/38.5 & 14.7/34.3 & 48.3/78.6 & 38.6/87.0 \\
& \hspace{0.5em}-ClipHigh-Ent      & 76.3/92.6 & 12.8/38.8 & 12.5/37.4 & 46.2/77.4 & 35.6/88.8 \\
& \hspace{0.5em}-Novelty Reward    & 79.3/88.7 & 12.1/27.0 & 11.1/34.8 & 47.6/73.3 & 37.9/81.4 \\
\midrule
\multirow{6}{*}{AIME24}
& \sert{\textbf{w/\textsc{Evol-RL}}}            & \sert{\textbf{79.6}/\textbf{93.6}} & \sert{\textbf{20.6}/\textbf{40.9}} & \sert{\textbf{17.1}/\textbf{42.0}} & \sert{\textbf{49.9}/\textbf{80.9}} & \sert{\textbf{38.0}/\textbf{87.8}} \\
& \hspace{0.5em}-ClipHigh          & 74.1/89.4 & 14.1/26.7 & 8.1/31.1  & 44.6/73.2 & 35.3/81.5 \\
& \hspace{0.5em}-Ent               & 66.7/89.8 & 10.0/31.4 & 6.6/27.8  & 38.7/74.2 & 34.0/86.2 \\
& \hspace{0.5em}-ClipHigh-Ent      & 75.3/89.0 & 16.6/26.9 & 9.2/32.2  & 45.8/71.2 & 37.1/82.0 \\
& \hspace{0.5em}-Novelty Reward    & 79.4/93.0 & 17.7/35.6 & 15.9/37.4 & 48.8/79.6 & 37.9/87.1 \\
\bottomrule
\end{tabular}
}
\vspace{-0.9em}
\end{table*}

\textbf{Setup.}
We ablate the three components of \textsc{Evol-RL}: novelty-based
within-group credit allocation, entropy regularization, and asymmetric
clipping. We remove each component individually, and remove the two supporting
mechanisms jointly, on Qwen3-4B-Base. Results are shown in
Table~\ref{tab:ablation}.

\textbf{Component roles depend on the training distribution.}
With MATH-500 training, removing novelty causes the largest degradation in
pass@16 on harder held-out tasks, reducing AIME24 from 43.2 to 27.0 and
AIME25 from 41.9 to 34.8. This indicates that within-group novelty is
particularly important when majority-based selection alone provides little
pressure to preserve alternative reasoning paths.

The pattern changes with AIME24 training. Here, removing entropy regularization
or asymmetric clipping causes larger drops than removing novelty on several
metrics. This is consistent with their supporting roles: entropy regularization
maintains a supply of varied rollouts, while asymmetric clipping preserves
strong updates from high-advantage samples. Overall, the ablation suggests that
the three components are complementary, with their relative importance
depending on the amount of variation already present in the training
distribution.


\begin{figure*}[t]
  \centering
  \includegraphics[width=0.875\linewidth]{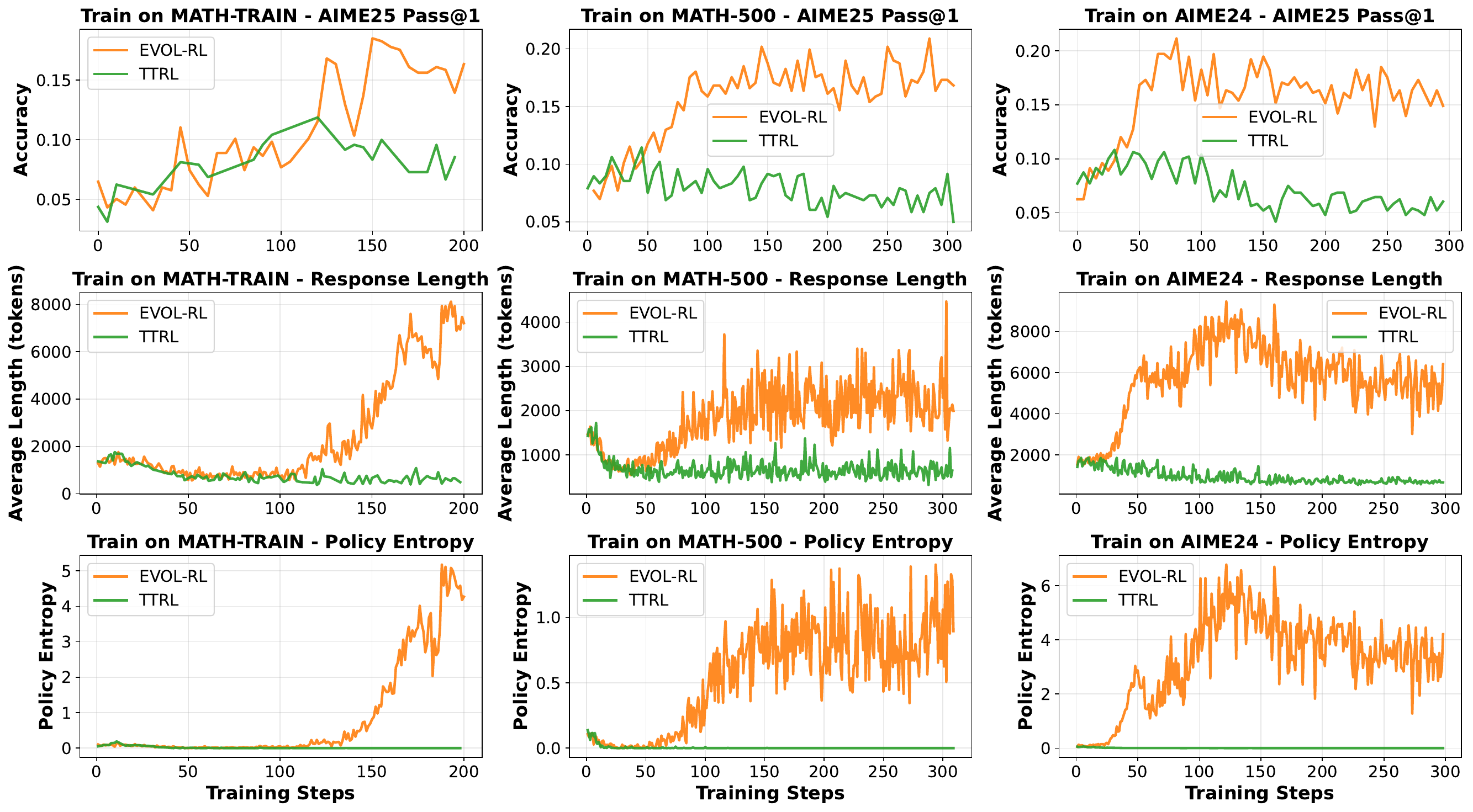}
  \caption{Training dynamics of TTRL and \textsc{Evol-RL}. Columns correspond
  to training on MATH-TRAIN, MATH-500, and AIME24; rows show held-out AIME25
  pass@1, response length, and policy entropy.}
  \vspace{-0.7em}
  \label{fig:dynamics}
\end{figure*}

\subsection{Training Dynamics: The Evolving Point and Recovery}
\label{sec:dynamics}

To understand how \textsc{Evol-RL} changes the training trajectory, we compare
its dynamics with TTRL in Figure~\ref{fig:dynamics}. Results for Qwen3-8B-Base
show the same overall pattern and are reported in
Appendix~\ref{app:8b_dynamics}.

\textbf{Stage 1: Initial Collapse Under Majority Signal.}
Across all three training sets, both methods initially show sharp decreases in
policy entropy and average response length. Their trajectories remain similar
during this phase, indicating that the majority-based selection signal initially
dominates the update and concentrates the policy. TTRL remains in this
low-entropy, short-response regime throughout the rest of training.

\textbf{Stage 2: The Evolving Point and Coordinated Recovery.}
\textsc{Evol-RL} subsequently diverges from TTRL at a clear
\emph{evolving point}, whose timing varies across training sets. After this
point, policy entropy and response length recover together with held-out
AIME25 pass@1, eventually stabilizing at higher levels, while TTRL remains
largely flat. Pass@16 trajectories show the same divergence
(Appendix~\ref{app:passn_curves}). The evolving point therefore marks the
transition from the initial selection-dominated phase to a recovery phase in
which broader solution coverage re-emerges.

This recovery is consistent with the complementary roles identified in the
ablation study: entropy regularization maintains candidate diversity, novelty
redistributes credit toward less redundant trajectories within each selection
group, and asymmetric clipping preserves stronger positive-advantage updates.

\subsection{What Does Variation Preserve?}
\label{sec:variation_preservation}

\begin{wraptable}{r}{0.4\linewidth}
    \centering
    \vspace{-0.8em}
    \renewcommand{\arraystretch}{1.12}
    \setlength{\tabcolsep}{3pt}
    \caption{Training dynamics from step 40 to 105 on Qwen3-4B-Base with MATH-500.}
    \vspace{-1mm}
    \resizebox{\linewidth}{!}{
    \begin{tabular}{lcc}
        \toprule
        & \textbf{TTRL} & \textbf{\textsc{Evol-RL}} \\
        \midrule
        Correct-minority
        & 45.1$\rightarrow$38.3
        & 45.3$\rightarrow$\textbf{55.5} \\
        MATH pass@16
        & 89.0$\rightarrow$88.5
        & 89.8$\rightarrow$\textbf{93.1} \\
        \bottomrule
    \end{tabular}}
    \label{tab:correct_minority}
\end{wraptable}

\textbf{Corrective alternatives.}
We first test whether variation keeps a correct answer reachable when the
current consensus is wrong. On Qwen3-4B-Base trained on MATH-500, we compare
step 40, before TTRL and \textsc{Evol-RL} begin to diverge, with step 105,
after the recovery emerges. We measure \emph{correct-minority availability}:
among prompts whose majority answer is incorrect, the fraction with at least
one correct minority response. As shown in Table~\ref{tab:correct_minority},
this availability falls from 45.1\% to 38.3\% under TTRL, but rises from
45.3\% to 55.5\% under \textsc{Evol-RL}, together with higher pass@16.
Thus, preserving variation helps keep corrective solutions reachable when the
current consensus is wrong. Full details are provided in
Appendix~\ref{app:correct_minority}.

\textbf{Distinct successful reasoning strategies.}
We next ask whether the preserved variation corresponds to different successful
approaches rather than merely diverse failures. On Qwen3-4B-Base trained on
AIME24, we evaluate three checkpoints---the base model, one before the evolving
point, and the final model---using 16 rollouts on the same 30 AIME24 problems.
Restricting to pairs in which both responses are correct,
Qwen3-235B-A22B-Instruct judges whether the two solutions use the same or
different mathematical approaches. The different-approach rate increases from
11.1\% in the base model to 22.2\% in the final model, while the number of
comparable correct-solution pairs grows from 25 to 51. This complements the
correct-minority analysis: \textsc{Evol-RL} preserves not only corrective
alternatives when consensus is wrong, but also multiple successful reasoning
paths. Full details are provided in Appendix~\ref{app:strategy_diversity}.

\subsection{\textsc{Evol-RL} Components Also Strengthen Supervised GRPO (RLVR)}

\begin{figure*}[t]
  \centering
  \includegraphics[width=0.88\linewidth]{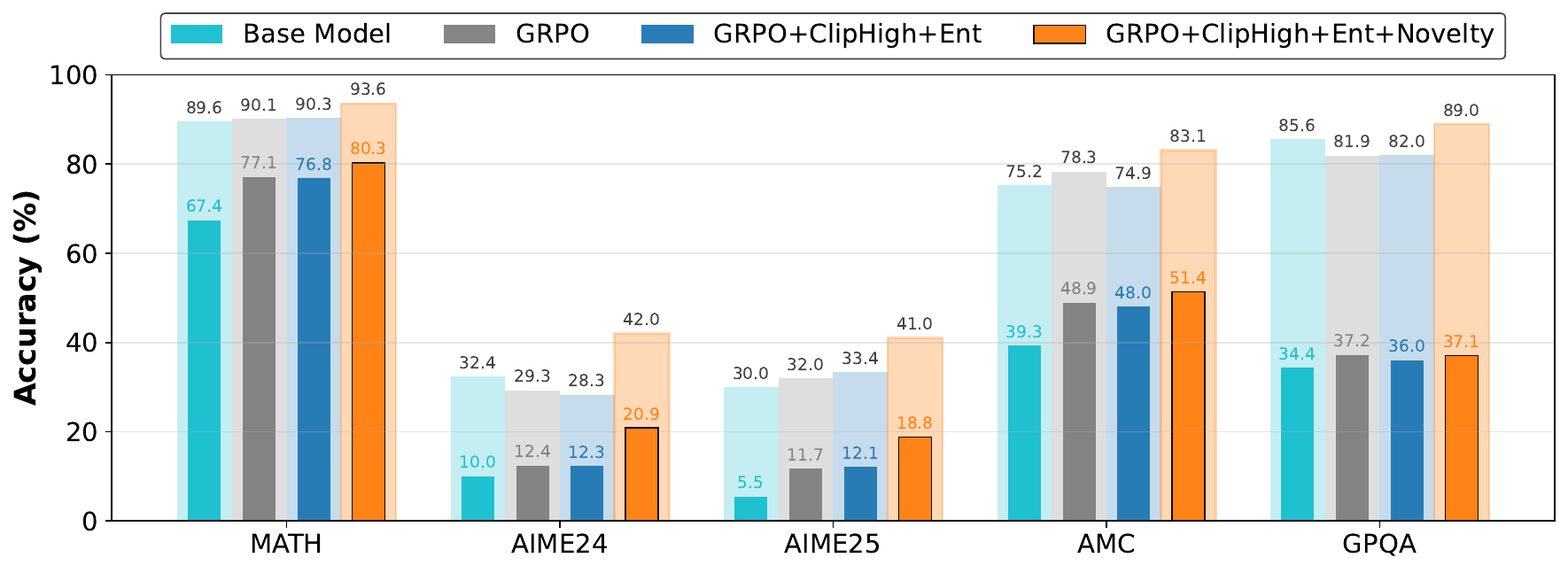}
  \vspace{-6pt}
  \caption{Performance of \textsc{Evol-RL} components applied to supervised
  GRPO. Qwen3-4B-Base is trained on MATH with a ground-truth verifier.
  Outer bars show pass@16 and inner bars show pass@1.}
  \label{fig:rlvr}
  \vspace{-0.7em}
\end{figure*}

\textbf{Setup.}
We apply the three components of \textsc{Evol-RL} to a standard supervised
GRPO baseline trained on the MATH training set
\citep{hendrycks2021measuring} with a ground-truth verifier for two epochs.
Unlike our main label-free setting, selection here is determined directly by
answer correctness rather than a self-generated majority signal.
Figure~\ref{fig:rlvr} reports the results.

The full combination, \texttt{GRPO+ClipHigh+Ent+Novelty}, achieves the strongest
overall performance. Compared with standard GRPO, it improves pass@16 by
12.7 points on AIME24 and 9.0 points on AIME25, while also improving pass@1
by 8.5 and 7.1 points, respectively. The same configuration improves pass@16
on MATH, AMC, and GPQA by 3.5, 4.8, and 7.1 points, respectively, while
improving or essentially maintaining pass@1. These results show that the
benefit of preserving variation is not limited to label-free majority-based
training: even when a ground-truth verifier provides a reliable selection
signal, maintaining diverse candidates and differentiating among correct
reasoning trajectories can further improve reasoning performance.

\begin{table}[t]
\centering
\renewcommand{\arraystretch}{1.22}
\caption{Transfer performance of the Qwen3-8B-Base model on broader reasoning benchmarks after label-free training on MATH-TRAIN.}
\label{tab:mmlu-gpqa-bbeh}
\resizebox{0.75\textwidth}{!}{
\begin{tabular}{lcccccc}
\toprule
\textbf{Model} & \multicolumn{2}{c}{\textbf{MMLU-Pro}} & \multicolumn{2}{c}{\textbf{SuperGPQA}} & \multicolumn{2}{c}{\textbf{BBEH}} \\
\cmidrule(lr){2-3} \cmidrule(lr){4-5} \cmidrule(lr){6-7}
& Pass@1 & Pass@4 & Pass@1 & Pass@4 & Pass@1 & Pass@4 \\
\midrule
Qwen3-8B-Base & 47.3 & 74.5 & 26.5 & 54.1 & 10.4 & 24.0 \\
~~~~~~w/TTRL        & 53.4 & 73.9 & 29.7 & 53.3 & \textbf{12.1} & 24.1 \\
\rowcolor{lightblue}
~~~~~~w/EVOL-RL     & \textbf{55.3} & \textbf{78.5} & \textbf{30.2} & \textbf{57.0} & 11.5 & \textbf{24.9} \\
\bottomrule
\end{tabular}
}
\vspace{-4mm}
\end{table}

\subsection{Transfer to Broader Reasoning Benchmarks}

To assess whether the improvements learned from mathematical data transfer to
broader reasoning tasks, we evaluate Qwen3-8B-Base after label-free training on
MATH-TRAIN on \textbf{MMLU-Pro} \citep{wang2024mmlupro},
\textbf{SuperGPQA} \citep{pteam2025supergpqascalingllmevaluation}, and
\textbf{BBEH} \citep{kazemi2025bigbenchextrahard}. None of these benchmarks is
used during training. The results are reported in
Table~\ref{tab:mmlu-gpqa-bbeh}.

A similar pass@1--pass@$k$ tension appears beyond the mathematical evaluation
sets. TTRL improves pass@1 substantially over the base model, but its pass@4
falls slightly below the base model on MMLU-Pro (73.9 vs.\ 74.5) and
SuperGPQA (53.3 vs.\ 54.1), indicating that stronger single-response
performance does not always translate into broader coverage. In contrast,
\textsc{Evol-RL} improves pass@4 over both TTRL and the base model on all three
benchmarks: 78.5 on MMLU-Pro, 57.0 on SuperGPQA, and 24.9 on BBEH.
Pass@1 also improves over TTRL on MMLU-Pro and SuperGPQA, while remaining
slightly lower on BBEH (11.5 vs.\ 12.1). These results extend the pattern
observed in our main experiments: preserving variation during label-free
mathematical training transfers positively to broader held-out reasoning tasks,
with the clearest gains appearing when multiple attempts are considered.

\section{Conclusion}
In this work, we identify \emph{selection-induced diversity collapse} as a
failure mode of label-free self-improvement: when supervision is repeatedly
constructed from a model's own outputs, selection can progressively narrow the
candidate pool and make alternative correct solutions harder to recover. We
propose \textsc{Evol-RL}, which addresses this problem by balancing
\emph{selection} and \emph{variation}: majority voting provides a stable
selection signal, while novelty-based credit allocation preserves variation
within each selection group, supported by entropy regularization and asymmetric
clipping. Across training settings, model scales, and matched self-improvement
baselines, \textsc{Evol-RL} improves reasoning performance while better
preserving solution coverage. Training analyses further show that it retains
correct alternatives when the current consensus is wrong, supporting the role
of variation in sustaining future opportunities for correction.

\bibliography{colm2024_conference}
\bibliographystyle{colm2024_conference}

\newpage
\appendix
\section{Implementation Details}
\label{app:implementation_details}

This section provides additional details on the implementation of our reward formulation and supporting mechanisms.

\subsection{Training Configuration.} 

Our main experiments use two open-source base models: \textbf{Qwen3-4B-Base} and \textbf{Qwen3-8B-Base}. Additional architecture and scale experiments with OctoThinker-8B-Hybrid-Base and Qwen3.5-35B-A3B-Base are reported in Appendix~\ref{sec:octothinker_results}. Our main training process is implemented using the GRPO algorithm. We adopt a setup similar to that of TTRL for generating training signals. For each problem instance, we first perform a rollout phase where the policy generates $64$ candidate responses. A majority label is then determined by performing a majority vote on the final answers extracted from these $64$ samples. Subsequently, a random subset of $32$ of these responses is used to form a batch for a single model update step. To ensure that the model has sufficient capacity for complex, multi-step reasoning, we set the maximum response length to 12,288 tokens during generation. To guide the model's reasoning process, we utilize the system prompt from SimpleRL-Zoo \citep{zeng2025simplerl}.

\subsection{System Prompt}
For the main experiments, we used the following system prompt to guide the model's generation format, ensuring that it produces a step-by-step reasoning process and a clearly marked final answer \citep{zeng2025simplerl}:
\begin{bluebox}[System Prompt]
\texttt{
Please reason step by step, and put your final answer within \textbackslash boxed\{\}.
}
\end{bluebox}

\subsection{Answer and Reasoning Extraction}
To implement the scoring criteria described in the main text, we apply the following extraction procedure for each generated response $o_i$:
\begin{itemize}
    \item \textbf{Final Answer Extraction (for Validity):} We parse the response to find the content within the final occurrence of the \verb|\boxed{·}| command. A response is deemed "valid" only if this command is present and its content contains at least one numeric digit. This extracted numeric string is used for the majority vote.
\end{itemize}

\subsection{Novelty Score Calculation Details}
\label{app:method_details}
The novelty score $u_i$ relies on computing semantic similarity between the reasoning parts of the generated responses.

\paragraph{Embedding Model.} We use the \textbf{Qwen3-Embedding-4B} model to generate dense vector representations for the extracted reasoning parts. Each vector is L2-normalized before similarity computation.

\paragraph{Cosine Similarity Matrix.} For a group of $G$ responses with corresponding L2-normalized embedding vectors $\{\mathbf{v}_1, \ldots, \mathbf{v}_G\}$, the cosine similarity matrix $\mathbf{S} \in \mathbb{R}^{G \times G}$ is computed as $\mathbf{S} = \mathbf{V}\mathbf{V}^T$, where $\mathbf{V}$ is the matrix whose rows are the vectors $\mathbf{v}_i$. The element $S_{ij}$ represents the cosine similarity between the reasoning of response $o_i$ and $o_j$.

\paragraph{Intra-Group Min-Max Normalization.} To obtain the normalized novelty score $\tilde{u}_i \in [0,1]$ from the raw scores $\{u_k\}$ within a specific group (e.g., the majority group), we apply standard min-max normalization:
\[
\tilde{u}_i = \frac{u_i - \min(\{u_k\})}{\max(\{u_k\}) - \min(\{u_k\}) + \epsilon_{\text{norm}}}
\]
where $\epsilon_{\text{norm}}$ is a small constant (e.g., $10^{-8}$) to prevent division by zero in cases where all novelty scores in the group are identical.

\subsection{Hyperparameter Settings}
For our main local label-free experiments, we largely follow the settings established by TTRL to ensure a fair comparison. The general hyperparameters are detailed in Table~\ref{tab:general_hyperparams}, and the settings specific to our \textsc{Evol-RL} method are listed in Table~\ref{tab:evolrl_hyperparams}.

\begin{table}[h]
\centering
\caption{General hyperparameters for label-free training, following TTRL.}
\label{tab:general_hyperparams}
\begin{tabular}{lc}
\toprule
\textbf{Hyperparameter} & \textbf{Value} \\
\midrule
Train Batch Size & 8 \\
PPO Mini-Batch Size & 1 (effective size of 32) \\
PPO Micro-Batch Size & 2 \\
Rollouts for Majority Vote & 64 \\
Rollouts Used for Training & 32 \\
Generation Temperature & 1.0 \\
Validation Temperature & 0.6 \\
Learning Rate & 5e-7 \\
Use KL Loss & True \\
KL Loss Coefficient & 0.001 \\
\bottomrule
\end{tabular}
\end{table}

\begin{table}[h]
\centering
\caption{Key hyperparameters specific to the \textsc{Evol-RL} framework.}
\label{tab:evolrl_hyperparams}
\begin{tabular}{lc}
\toprule
\textbf{Hyperparameter} & \textbf{Value} \\
\midrule
Asymmetric Clipping High ($\epsilon_{\text{high}}$) & 0.28 \\
Entropy Regularizer Coefficient ($\lambda_{\text{ent}}$) & 0.003 \\
Novelty Score Mixing Coefficient ($\alpha$) & 0.5 \\
\bottomrule
\end{tabular}
\end{table}

\subsection{Computational Resources}
The main local experiments were conducted on a server with 8 NVIDIA H20 GPUs.
The Qwen3.5-35B-A3B-Base scale experiment was run through Tinker using LoRA;
its configuration and results are reported in
Appendix~\ref{sec:octothinker_results}.

\section{Additional Experimental Results}
\label{app:additional_experimental_results}

{
\FloatBarrier
\subsection{Effectiveness Across Model Architectures and Scale}
\label{sec:octothinker_results}

We further examine whether the gains of \textsc{Evol-RL} extend beyond the
Qwen3-4B/8B configurations used in our main experiments along two complementary
dimensions: model architecture and model scale. We first evaluate a different
8B architecture, OctoThinker-8B-Hybrid-Base, and then conduct a short-horizon
scale check on Qwen3.5-35B-A3B-Base.

\paragraph{OctoThinker-8B-Hybrid-Base.}
To test whether the observed behavior is specific to the Qwen3 architecture,
we apply \textsc{Evol-RL} and TTRL to
\textbf{OctoThinker-8B-Hybrid-Base} \citep{wang2025octothinker}, using the
same label-free training setup on MATH-500. As shown in
Table~\ref{tab:octothinker}, TTRL substantially improves MATH pass@1
(from 33.8\% to 63.8\%), but its pass@16 decreases relative to the base model
on AIME24 (13.5\% to 10.8\%) and GPQA (85.7\% to 71.5\%). In contrast,
\textsc{Evol-RL} achieves comparable MATH pass@1 while improving pass@16 over
TTRL across all five benchmarks, including gains of 19.5 points on AIME24 and
11.4 points on AIME25.

\begin{table}[!htbp]
\centering
\renewcommand{\arraystretch}{1.25}
\setlength{\tabcolsep}{6pt}
\caption{Comparison of models trained with TTRL and \textsc{Evol-RL} on MATH-500 on OctoThinker-8B-Hybrid-Base. Each cell shows pass@1/pass@16 (averaged over 32 rollouts). $\Delta$ uses red (+) for positive and blue for negative values.}
\label{tab:octothinker}
\vspace{-0.3em}
\resizebox{0.74\columnwidth}{!}{
\begin{tabular}{cccccc}
\toprule
\textbf{Model} & \textbf{MATH} & \textbf{AIME24} & \textbf{AIME25} & \textbf{AMC} & \textbf{GPQA} \\
\rowcolor{lightgray}
\midrule
\multicolumn{6}{c}{\textbf{OctoThinker-8B-Hybrid-Base}} \\
\midrule

Base Model 
& 33.8/79.8 & 1.5/13.5 & 1.3/13.9 & 16.2/56.9 & 26.3/85.7 \\
\midrule

w/TTRL
& 63.8/76.4 & 2.8/10.8 & 2.1/11.0 & 27.9/54.7 & 31.9/71.5 \\

\sert{w/\textsc{Evol-RL}}
& \sert{63.2/86.3} & \sert{9.0/30.3} & \sert{7.2/22.4} & \sert{34.1/65.6} & \sert{33.2/85.7} \\

$\Delta$
& \negcolor{-0.6}/\pos{9.9} & \pos{6.2}/\pos{19.5} & \pos{5.1}/\pos{11.4} 
& \pos{6.2}/\pos{10.9} & \pos{1.3}/\pos{14.2} \\

\bottomrule
\end{tabular}
}
\vspace{-0.5em}
\end{table}
\FloatBarrier

\begin{samepage}
\paragraph{Qwen3.5-35B-A3B-Base.}
We next examine whether the same pattern extends beyond the model sizes used in
our main experiments. We train Qwen3.5-35B-A3B-Base \citep{qwen2026qwen35} on AIME24 using LoRA
through Tinker \citep{thinkingmachines2025tinker}, with both TTRL and \textsc{Evol-RL} trained for 30 steps.
Because of this limited training horizon, we treat the experiment as a scale
check rather than a fully converged comparison. As shown in
Table~\ref{tab:qwen35b}, both methods improve over the base model on AIME24,
but \textsc{Evol-RL} achieves higher pass@1 and pass@16 than TTRL on both
AIME24 and held-out AIME25. The difference is again clearer in multi-sample
coverage, with pass@16 improving from 86.2 to 88.4 on AIME24 and from 67.9 to
71.4 on AIME25.

\end{samepage}
\begin{table}[!htbp]
\centering
\renewcommand{\arraystretch}{1.25}
\setlength{\tabcolsep}{6pt}
\caption{Short-horizon comparison of TTRL and \textsc{Evol-RL} on Qwen3.5-35B-A3B-Base trained on AIME24 for 30 steps using LoRA. Each cell shows pass@1/pass@16. $\Delta$ uses red (+) for positive and blue for negative values.}
\label{tab:qwen35b}
\vspace{-0.3em}
\resizebox{0.38\columnwidth}{!}{
\begin{tabular}{ccc}
\toprule
\textbf{Model} & \textbf{AIME24} & \textbf{AIME25} \\
\rowcolor{lightgray}
\midrule
\multicolumn{3}{c}{\textbf{Qwen3.5-35B-A3B-Base}} \\
\midrule

Base Model
& 57.0/80.8 & 44.9/67.6 \\
\midrule

w/TTRL
& 63.0/86.2 & 45.3/67.9 \\

\sert{w/\textsc{Evol-RL}}
& \sert{63.6/88.4} & \sert{48.2/71.4} \\

$\Delta$
& \pos{0.6}/\pos{2.2}
& \pos{2.9}/\pos{3.5} \\

\bottomrule
\end{tabular}
}
\vspace{-0.5em}
\end{table}
\FloatBarrier

Together, these results provide complementary evidence that the improvements
of \textsc{Evol-RL} are not confined to the particular architecture or model
sizes used in our main Qwen3-4B/8B experiments.

\FloatBarrier
\subsection{Additional Self-Improvement Baselines}
\label{sec:comparison_other_baselines}

The main text compares \textsc{Evol-RL} with TTRL, EMPO, Co-Reward, and ECHO
under a matched MATH-500 protocol. Here, we provide additional matched
comparisons on MATH-TRAIN against Self-Consistency
\citep{wang2022self,huang2023large} and the confidence-based
EM-RL-Token and EM-RL-Sequence objectives
\citep{agarwal2025unreasonable}. All methods use Qwen3-4B-Base and the same
label-free training and evaluation protocol. Results are reported in
Table~\ref{tab:more_baselines}.

\begin{table}[!htbp]
\centering
\renewcommand{\arraystretch}{1.25}
\setlength{\tabcolsep}{6pt}
\caption{Extended comparison of baseline methods on Qwen3-4B-Base, trained in a label-free setting using MATH-Train. Each cell shows pass@1/pass@16, and the highest value in each column is bolded.}
\label{tab:more_baselines}
\vspace{-0.3em}

\resizebox{0.74\columnwidth}{!}{
\begin{tabular}{lcccccc}
\toprule
\textbf{Model} & \textbf{MATH} & \textbf{AIME24} & \textbf{AIME25} & \textbf{AMC} & \textbf{GPQA} \\
\rowcolor{lightgray}
\midrule
\rowcolor{lightgray}
\multicolumn{6}{c}{\textbf{Qwen3-4B-Base}} \\
\midrule

Base Model
& 67.4/89.6 & 10.0/32.4 & 5.5/30.0 & 39.3/75.2 & 34.4/85.6 \\
\midrule

w/TTRL
& 75.4/86.9 & 12.1/23.2 & 6.8/28.6 & 42.5/75.2 & 36.5/81.4 \\

w/EM-RL-Token
& 76.0/90.6 & 12.5/31.3 & 10.5/30.8 & 46.6/77.7 & 36.8/82.6 \\

w/EM-RL-Sequence
& 67.4/89.9 & 10.6/31.4 & 7.1/28.8 & 39.6/73.7 & 34.5/86.2 \\

w/Self-Consistency
& 76.0/89.7 & 12.5/30.4 & 10.4/33.6 & 48.1/78.1 & 35.9/81.2 \\

\rowcolor{blue!10}
\textbf{w/\textsc{Evol-RL}}
& \textbf{80.0}/\textbf{93.3}
& \textbf{20.7}/\textbf{47.6}
& \textbf{17.5}/\textbf{39.9}
& \textbf{51.4}/\textbf{80.3}
& \textbf{37.2}/\textbf{88.7} \\

\bottomrule
\end{tabular}
}
\vspace{-0.5em}
\end{table}

\paragraph{Baseline implementation.}
We use the official TTRL implementation with the shared backbone and a maximum
response length of 12,288 tokens. EMPO retains the semantic-entropy reward
from its official verl implementation, and Co-Reward retains its defining
contrastive-agreement reward within the shared training framework. For ECHO,
we follow the authors' released MATH reproduction configuration, using the
official policy-update components and their method-specific default
hyperparameters.
These comparisons share the backbone, training prompts, rollout budget, and
evaluation protocol, while preserving method-specific reward and update
settings; they are not reproductions of each independently tuned pipeline.

A similar pattern emerges. TTRL, EM-RL, and Self-Consistency improve pass@1
on several benchmarks, but their pass@16 gains are less consistent. On AIME24,
for example, all four alternatives remain below the base model in pass@16
(23.2--31.4 vs.\ 32.4), whereas \textsc{Evol-RL} reaches 47.6. On AIME25,
\textsc{Evol-RL} achieves 17.5/39.9 in pass@1/pass@16, compared with
10.5/30.8 for EM-RL-Token and 10.4/33.6 for Self-Consistency. These additional
comparisons reinforce the trend observed in the main text: improving the
self-generated training signal does not necessarily preserve solution coverage,
while explicitly maintaining variation improves both accuracy and coverage.

\FloatBarrier
\subsection{Correct-Minority Availability}
\label{app:correct_minority}

We further examine whether correct alternatives remain reachable when the
current majority answer is wrong. At checkpoint $t$, we define
\emph{correct-minority availability} as
\[
\frac{
\#\{\text{prompts with an incorrect majority and at least one correct minority response}\}
}{
\#\{\text{prompts with an incorrect majority}\}
}.
\]
The denominator is recomputed independently at each checkpoint.

We analyze Qwen3-4B-Base trained on MATH-500. Step 40 is chosen as a
pre-divergence checkpoint: TTRL and \textsc{Evol-RL} still have nearly
identical entropy and similar accuracy and
correct-minority availability (Figure \ref{fig:dynamics}). By step 105, their training trajectories have
clearly diverged.

\begin{table}[t]
\centering
\small
\setlength{\tabcolsep}{5pt}
\caption{Training dynamics from step 40 to 105 on Qwen3-4B-Base trained on
MATH-500.}
\label{tab:correct_minority_full}
\begin{tabular}{lcc}
\toprule
\textbf{Metric} & \textbf{TTRL} & \textbf{\textsc{Evol-RL}} \\
\midrule
MATH pass@1 & 75.4 $\rightarrow$ 77.8 & 76.1 $\rightarrow$ 78.6 \\
MATH pass@16 & 89.0 $\rightarrow$ 88.5 & 89.8 $\rightarrow$ 93.1 \\
Correct-minority availability & 45.1 $\rightarrow$ 38.3 & 45.3 $\rightarrow$ 55.5 \\
Mean response length & 524 $\rightarrow$ 488 & 593 $\rightarrow$ 1,983 \\
\bottomrule
\end{tabular}
\end{table}

The two methods exhibit different trajectories. Under TTRL, pass@1 improves
slightly, but correct-minority availability, pass@16, entropy, and response
length all decrease. Thus, majority-only training can strengthen the current
prediction while making corrective alternatives less likely to remain sampled.
Under \textsc{Evol-RL}, correct-minority availability instead rises together
with pass@16, entropy, and response length.

The same pattern persists over the full run. From step 0 to step 180,
correct-minority availability decreases from 46.6\% to 34.4\% under TTRL,
while remaining roughly stable under \textsc{Evol-RL} (50.1\% to 50.9\%).
Over the same interval, MATH pass@16 changes from 88.3\% to 88.0\% for TTRL
and from 89.3\% to 93.9\% for \textsc{Evol-RL}. These results indicate that
\textsc{Evol-RL} preserves access to correct alternatives when the
self-generated majority is wrong, although it does not guarantee that an
incorrect majority will eventually be overturned.

\FloatBarrier
\subsection{Distinct Successful Reasoning Strategies}
\label{app:strategy_diversity}

We further examine whether the variation preserved by \textsc{Evol-RL}
corresponds to genuinely different successful reasoning strategies, rather than
merely diverse outputs or failure modes. We analyze Qwen3-4B-Base trained on
AIME24 at three checkpoints: the base model, one checkpoint before the evolving
point, and the final model. At each checkpoint, we generate 16 rollouts for the
same 30 AIME24 problems.

We restrict the analysis to pairs in which both solutions are correct. This
avoids conflating reasoning diversity with diversity among incorrect solutions:
at the base checkpoint, most generated responses are incorrect, so evaluating
all pairs would largely measure different failure modes rather than variation
among successful reasoning strategies.

For each eligible pair, Qwen3-235B-A22B-Instruct judges whether the two
solutions use the same or different mathematical approaches. The judge receives
only the problem and two anonymized solutions, labeled Solution A and
Solution B, without information about their checkpoint of origin. Each pair is judged twice with the solution order reversed, and we retain only
pairs for which the two judgments agree. Results are reported in
Table~\ref{tab:strategy_diversity}.

\begin{table}[!htbp]
\centering
\small
\setlength{\tabcolsep}{6pt}
\caption{Reasoning-strategy diversity across training checkpoints on AIME24.
The analysis includes only pairs in which both solutions are correct.}
\label{tab:strategy_diversity}
\begin{tabular}{lcc}
\toprule
\textbf{Checkpoint} &
\textbf{Different-Approach Rate} &
\textbf{Correct Pairs} \\
\midrule
Base                  & 11.1\% & 25 \\
Before evolving point & 14.3\% & 45 \\
Final                 & 22.2\% & 51 \\
\bottomrule
\end{tabular}
\end{table}
\FloatBarrier

Among correct solutions, the fraction judged to use different mathematical
approaches increases from 11.1\% for the base model to 22.2\% for the final
model. The number of comparable correct-solution pairs also increases from 25
to 51 as accuracy on these problems rises from 8.5\% to 18.8\%. Thus, the
final model not only produces more correct solutions, but its correct solutions
are also more likely to represent distinct reasoning approaches.

Because the number of correct-solution pairs is limited, particularly for the
base checkpoint, we treat this analysis as direct but exploratory evidence.
Together with the correct-minority analysis in
Appendix~\ref{app:correct_minority}, it suggests that \textsc{Evol-RL}
preserves both corrective alternatives when the current consensus is wrong and
multiple successful reasoning paths when correct solutions are available.

\FloatBarrier
\subsection{Analysis of the Majority-Vote Signal}
\label{app:maj_vote_signal}

To examine how preserving variation affects the self-generated supervision
available later in training, we track the accuracy of the majority-voted answer
over time. At each checkpoint, we derive a consensus answer from 16 rollouts
and evaluate it against the ground-truth answer on AIME25; ground truth is used
only for this analysis and is never provided during label-free training.
Figure~\ref{fig:maj16_dynamics} reports the resulting maj@16 trajectories.

\begin{figure}[!htbp]
  \centering
  \includegraphics[width=1\linewidth]{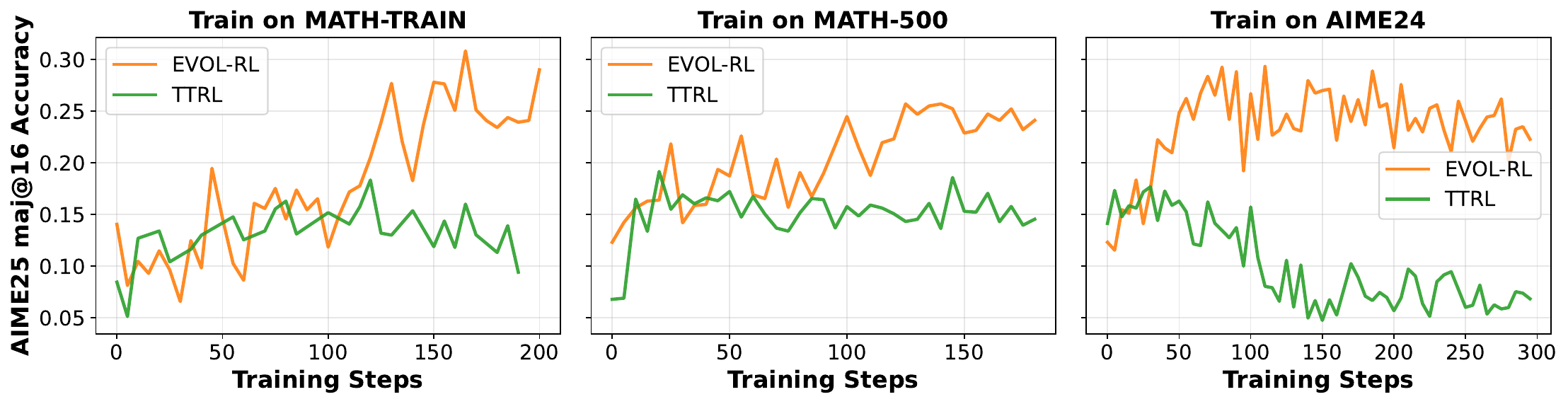} 
 \caption{Majority-vote accuracy (maj@16) on AIME25 throughout training.
The models are trained on (\textbf{Left}) MATH-TRAIN,
(\textbf{Middle}) MATH-500, and (\textbf{Right}) AIME24.}
  \label{fig:maj16_dynamics}
  \vspace{-6pt}
\end{figure}

Across all three training distributions, TTRL and \textsc{Evol-RL} initially
follow similar trajectories. TTRL's majority-vote accuracy improves early but
then largely plateaus. In contrast, \textsc{Evol-RL} subsequently diverges near
the evolving point and continues to improve its majority-vote accuracy,
eventually stabilizing at a higher level.

Because the majority-voted answer is itself used to construct the training
signal, this difference is important for the self-improvement loop. Preserving
variation does not merely maintain alternative responses at the current step;
it also leaves a broader candidate pool from which later majority signals are
constructed. The continued increase in maj@16 under \textsc{Evol-RL} is
therefore consistent with progressively improving the quality of its future
self-generated supervision, whereas TTRL's supervision quality changes little
after its early improvement.

\FloatBarrier
\subsection{Training Dynamics of 8B Models}
\label{app:8b_dynamics}

We further examine whether the training dynamics observed with
Qwen3-4B-Base persist at a larger model scale. Figure~\ref{fig:8b_dynamics}
reports AIME25 pass@1, average response length, and policy entropy throughout
training for Qwen3-8B-Base across MATH-TRAIN, MATH-500, and AIME24.

Across all three training distributions, the 8B model exhibits the same
two-stage pattern observed in the 4B experiments. During the initial stage,
TTRL and \textsc{Evol-RL} follow similar trajectories: policy entropy and
response length decrease as majority-based selection concentrates the policy.
TTRL largely remains in this concentrated regime for the rest of training.

\begin{figure}[t]
  \centering
  \includegraphics[width=\linewidth]{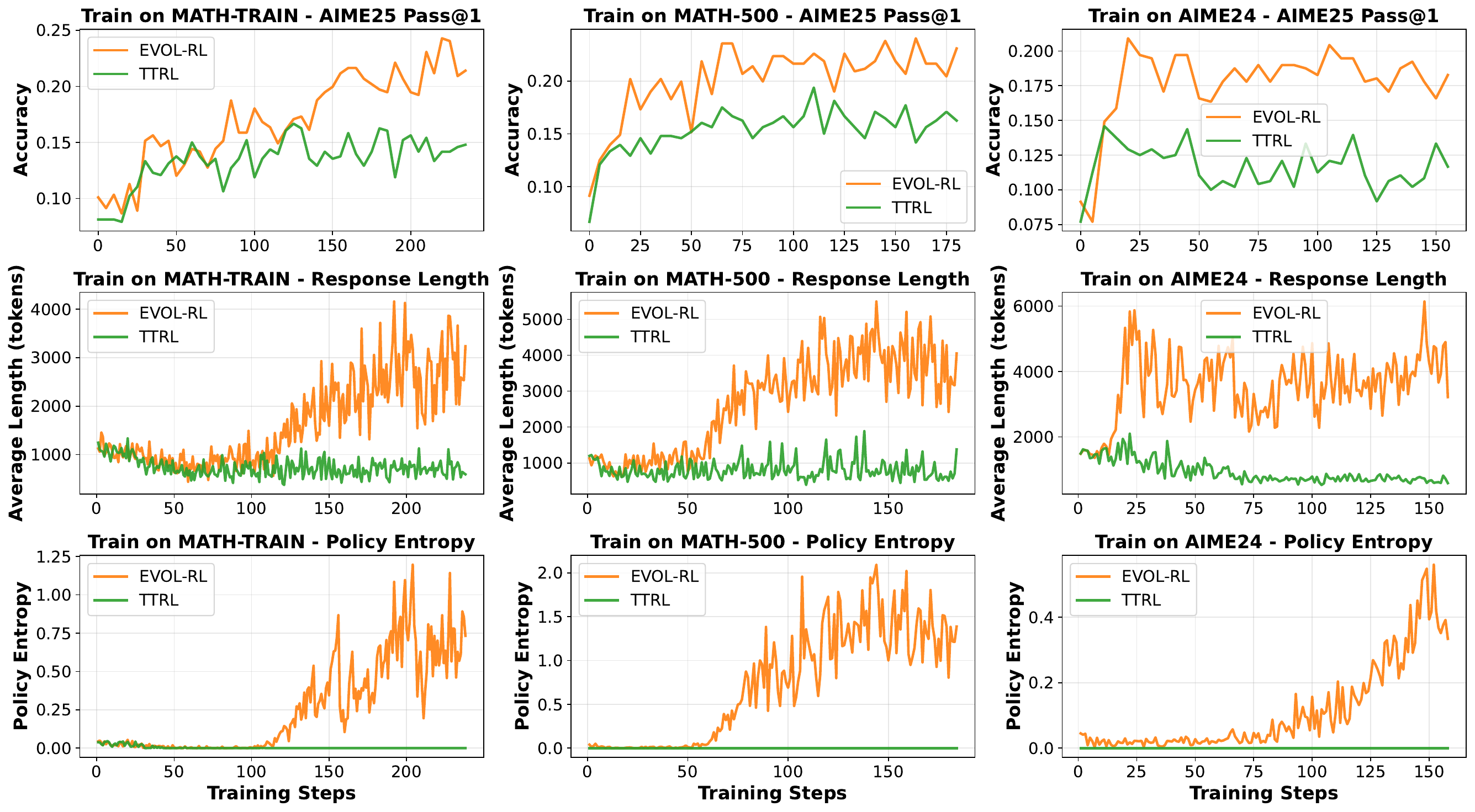}
  \caption{Training dynamics of TTRL and \textsc{Evol-RL} on
  Qwen3-8B-Base. Columns correspond to training on MATH-TRAIN, MATH-500,
  and AIME24; rows show held-out AIME25 pass@1, average response length,
  and policy entropy.}
  \label{fig:8b_dynamics}
\end{figure}

\textsc{Evol-RL}, however, subsequently diverges at an evolving point. After
this point, policy entropy and response length recover together with held-out
AIME25 pass@1, while the corresponding TTRL trajectories remain substantially
flatter. The timing of the evolving point differs across training
distributions, but the qualitative pattern is consistent. These results show
that the recovery dynamics observed in the main 4B experiments are not
specific to a single model scale.

\subsection{Pass@16 Training Curves}
\label{app:passn_curves}

To complement the training dynamics in Section~\ref{sec:dynamics},
Figure~\ref{fig:passn_curves} reports the pass@16 trajectories on AIME24
and AIME25 across all three training settings. TTRL's pass@16 generally
declines or plateaus over training, whereas \textsc{Evol-RL}'s pass@16
recovers after the evolving point and stabilizes at a higher level. This
provides additional evidence that the recovery observed in policy entropy and
response length is accompanied by broader solution coverage.

\begin{figure*}[t]
  \centering
  \includegraphics[width=1\linewidth]{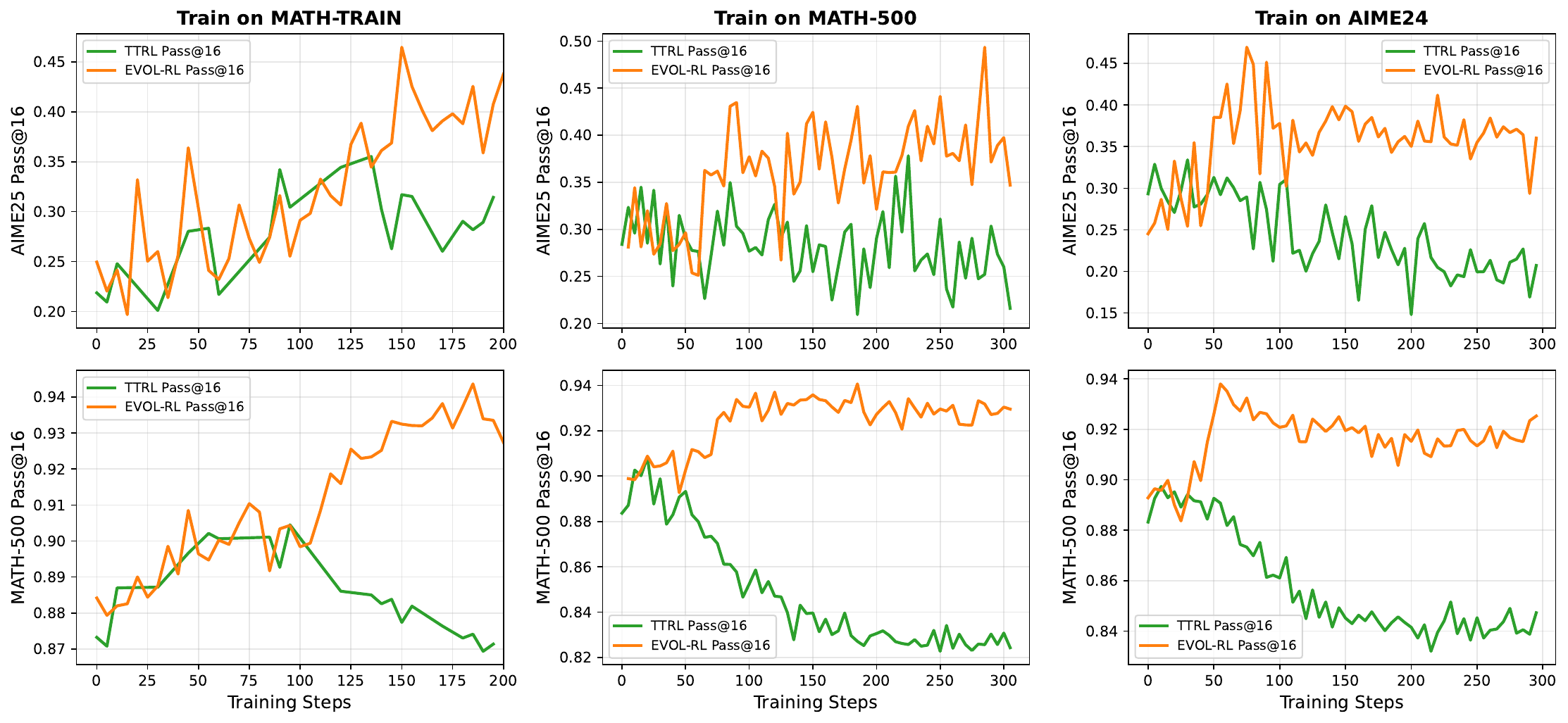}
  \caption{Pass@16 training curves on AIME24 and AIME25 across three training settings (MATH-TRAIN / MATH-500 / AIME24). TTRL's pass@16 consistently declines, while \textsc{Evol-RL}'s steadily recovers and grows after the evolving point.}
  \label{fig:passn_curves}
\end{figure*}

\subsection{Analysis of the Computational Overhead from the Novelty Reward}
\label{app:compute_overhead}

\begin{figure}[t]
    \centering
    \includegraphics[width=1\textwidth]{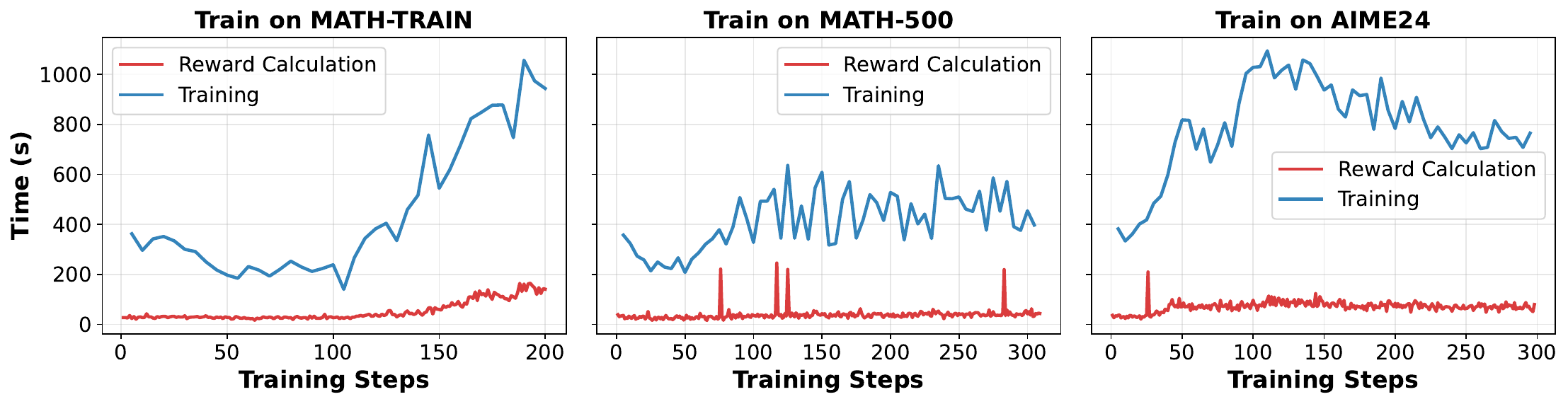} 
    \caption{Wall-clock time (seconds) per training step for \textbf{Novelty Reward Calculation} (Red) versus the \textbf{Total Training Process} (Blue) across three datasets. While both costs naturally increase as the model evolves to generate longer reasoning paths, the reward calculation overhead remains a minor fraction of the total computational load.}
    \label{fig:compute_overhead}
\end{figure}

A practical consideration for our method is the additional computational load introduced by the embedding-based novelty reward. To rigorously quantify this, we tracked the wall-clock time per training step, decomposing it into two parts: (1) the {Novelty Reward Calculation} (comprising $B \times N$ embedding model calls and the $O(B N^2)$ similarity matrix computation, where $B$ is the batch size and $N$ is the number of rollouts), and (2) the Total {Training Process} (including rollout generation, reward calculation and model updates). The results are plotted in Figure~\ref{fig:compute_overhead}.

The analysis reveals two key insights:
\begin{itemize}
    \item \textbf{Scaling with Reasoning Length:} As expected, both curves rise over time. This correlates with our findings in Figure \ref{fig:dynamics} that the model learns to generate significantly longer CoT during training. Longer responses require more time to generate (Training) and more time to embed (Reward Calculation).
    \item \textbf{Low Relative Overhead:} while the reward cost scales with response length (requiring embedding of longer sequences), the Novelty Reward Calculation (red curve) remains a small portion of the overall runtime, stabilizing around 100 seconds.
\end{itemize}

The occasional sharp spikes in the Red curve are anomalous and likely attributable to external API connection instabilities. We conclude that the embedding-based reward introduces a manageable overhead
relative to the overall cost of training the reasoning model.

\subsection{Case Study: Validity of the Semantic Novelty Proxy}
\label{app:case_study_similarity}

A potential concern regarding our method is whether semantic similarity in an embedding space is a meaningful proxy for genuine reasoning novelty. To validate this, we conducted a case study, shown in Figure~\ref{fig:similarity_case_study}, by generating 8 rollouts for a single AIME25 problem using Qwen3-4B-Instruct-2407 and computing their reasoning path similarity matrix with Qwen3-Embedding-4B.

The results provide strong empirical evidence for the validity of this proxy. As shown in the figure, the matrix reveals distinct, high-similarity clusters that align with the reasoning logic. Specifically, the 6 rollouts that produced the same final answer ("19") have an extremely high intra-cluster similarity (the 6x6 block at the top-left). Crucially, these paths are semantically distinct from the path that answered "31" and the path that answered "7221", showing clear separation in the embedding space.

This demonstrates that the embedding space does successfully capture and cluster the core logic of the reasoning paths, distinguishing between different reasoning trajectories.

\begin{figure}[t]
    \centering
    \includegraphics[width=0.6\textwidth]{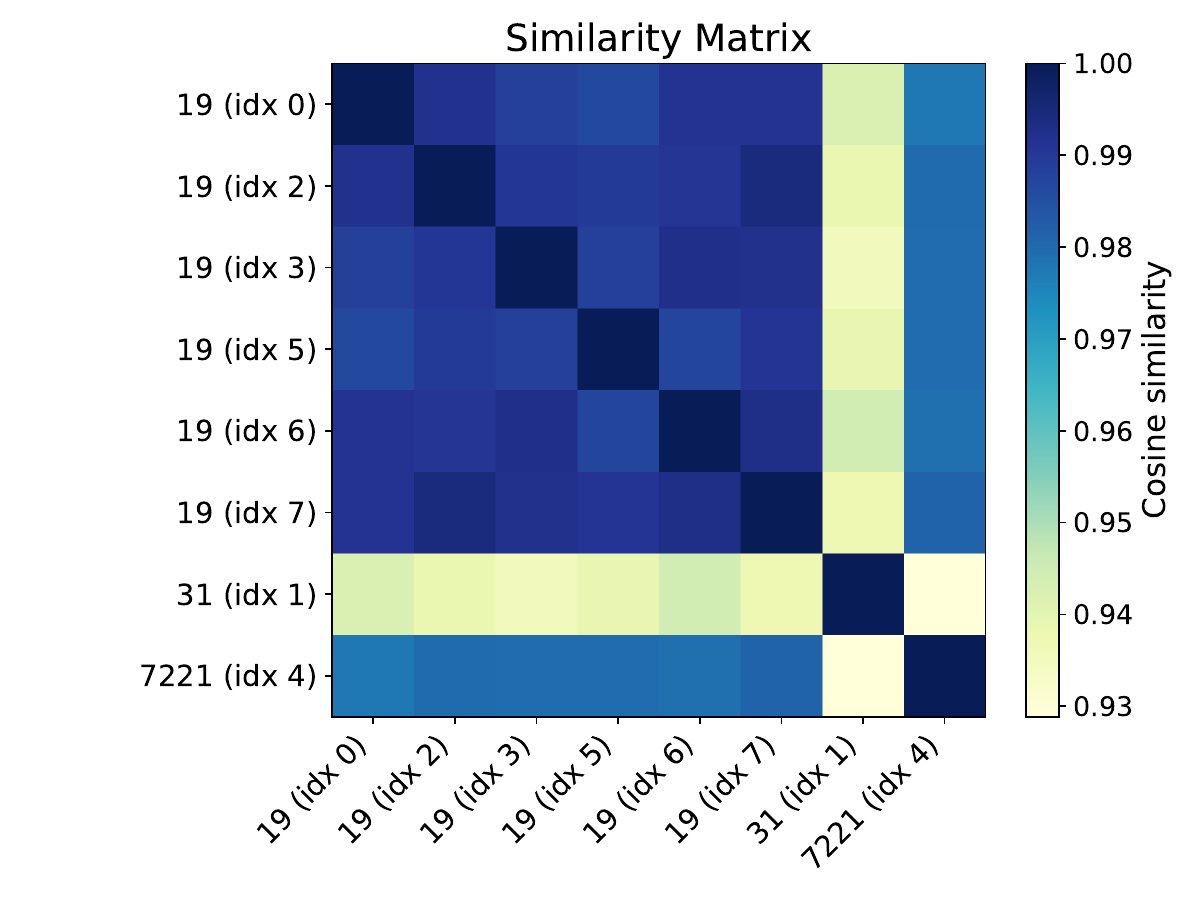}
    \caption{
        Case study of the reasoning path cosine similarity matrix for 8 rollouts on a single AIME25 problem. The axes are labeled with the \texttt{final\_answer(index\_i)} for each rollout. The 8 rollouts produced three distinct final answers: "19" (6 times), "31" (1 time), and "7221" (1 time).
    }
    \label{fig:similarity_case_study}
\end{figure}

\subsection{LLM-Judge Validation of the Cosine Similarity Proxy}
\label{app:llm_judge_validation}

We validate that cosine similarity in the embedding space tracks semantic reasoning differences rather than surface artifacts such as length, formatting, or phrasing. To this end, we ask an independent LLM judge---DeepSeek-V3.2, drawn from a different model family than the embedding---to label whether each pair of rollouts shares the same mathematical approach. The evaluation pool consists of 60 AIME questions with 64 \textsc{Evol-RL} rollouts each (Qwen3-4B-Base, MATH-500 training); for each question we sample pairs from the top-third (high-similarity) and bottom-third (low-similarity) of its cosine distribution and submit them anonymously to the judge.

Across the 60 questions, the judge agrees with the cosine ranking on 83.9\% of sampled pairs. The agreement is asymmetric in a way that supports the design: high-similarity pairs are labeled SAME at 77.2\%, while low-similarity pairs are labeled DIFFERENT at 90.6\%---i.e., the embedding rarely flags two semantically distinct solutions as similar. The mean cosine gap between SAME and DIFFERENT pairs is $+0.436$, well above any plausible noise floor.

We further verify that the proxy remains discriminative on long reasoning traces, where all embeddings could in principle be pushed toward a common region. Table~\ref{tab:llm_judge_length} bins the questions by response length and shows the opposite of any such collapse: agreement rises with length, from 75.0\% on the shortest bin ($<$3K tokens) to 92.2\% on the longest ($>$6K tokens), and the SAME/DIFFERENT cosine gap widens correspondingly. Longer traces give the embedding more structural signal to work with, not less.

\begin{table}[t]
\centering
\small
\caption{LLM-judge agreement with cosine-similarity ranking, grouped by response length.}
\label{tab:llm_judge_length}
\begin{tabular}{lccccc}
\toprule
Length & \# Q & High-sim$\to$SAME & Low-sim$\to$DIFF & Agreement & Sim. Gap \\
\midrule
$<3$K & 12 & $66.7\%$ & $83.3\%$ & $75.0\%$ & $+0.301$ \\
3K--4.5K & 20 & $80.0\%$ & $91.7\%$ & $85.8\%$ & $+0.466$ \\
4.5K--6K & 13 & $69.2\%$ & $89.7\%$ & $79.5\%$ & $+0.378$ \\
6K+ & 15 & $88.9\%$ & $95.6\%$ & $92.2\%$ & $+0.550$ \\
\midrule
\textbf{Total} & \textbf{60} & $\mathbf{77.2\%}$ & $\mathbf{90.6\%}$ & $\mathbf{83.9\%}$ & $\mathbf{+0.436}$ \\
\bottomrule
\end{tabular}
\end{table}

\subsection{Surface-Rewriting Robustness}
\label{app:surface_rewrite}

We next test whether surface-level paraphrasing can inflate the novelty signal: an agent that simply restates the same solution in different language should not be rewarded as ``novel'' since no real reasoning variation has occurred. We use DeepSeek-V3.2 to surface-rewrite 300 solutions (30 AIME24/25 questions $\times$ 5 solutions each), with explicit instructions to preserve every reasoning step while polishing language and notation.

Across the 300 refine pairs, the cosine similarity between a refined solution and its original is 0.834, well above the average similarity of 0.681 to unrelated solutions on the same question (Table~\ref{tab:surface_rewrite}). The gap of $+0.154$ between within-pair and cross-pair similarity is highly significant ($p < 0.001$). Surface rewriting therefore moves a solution only marginally in embedding space, whereas genuinely distinct reasoning paths produce substantially larger displacement. The novelty signal is sensitive to the underlying reasoning structure, not the surface form.

\begin{table}[t]
\centering
\small
\caption{Surface-rewriting robustness on $300$ solutions ($30$ AIME24/25 questions $\times$ $5$ each). Refined solutions remain close to their originals, while cross-solution pairs are clearly more distant.}
\label{tab:surface_rewrite}
\begin{tabular}{lc}
\toprule
Comparison & Cosine Similarity \\
\midrule
Refined vs.\ original (within-pair) & $0.834$ \\
Cross-solution (different reasoning) & $0.681$ \\
\midrule
\textbf{Gap} & $\mathbf{+0.154}$ \\
\bottomrule
\end{tabular}
\end{table}

\FloatBarrier
\subsection{Cross-Family Embedding Robustness}
\label{app:cross_family_embedding}

We finally examine the framework's sensitivity to the embedding model family. Because the policy is Qwen-based, one might wonder whether the Qwen-family embedding contributes inductive biases that another embedding would not, so we re-run the full pipeline after replacing Qwen3-Embedding-4B with OpenAI's text-embedding-3-small---an entirely different model family with a different training corpus and architecture---while keeping every other configuration fixed.

\begin{samepage}
Both embedding choices improve \texttt{pass@16} over the no-novelty ablation across MATH, AIME24, and AIME25, while maintaining or improving \texttt{pass@1}. On the most challenging benchmark, AIME24 \texttt{pass@16} reaches 43.2\% with the Qwen embedding and 39.7\% with the OpenAI embedding---both far above the 27.0\% obtained without any novelty signal. The MATH numbers are essentially identical (93.8\% with either embedding vs.\ 88.7\% without), and AIME25 follows the same pattern (Table~\ref{tab:cross_family_embedding}).\par
\end{samepage}

The 3.5-pp residual gap between the two embeddings on AIME24 \texttt{pass@16} is consistent with the OpenAI model's 8K-token context limit, which truncates the longer reasoning traces that carry the most distinctive structure. Once that artifact is accounted for, the framework is essentially indifferent to the embedding family.

\begin{table}[!htbp]
\centering
\small
\caption{Cross-family embedding ablation; both embeddings improve multi-sample coverage over the no-novelty ablation.}
\label{tab:cross_family_embedding}
\begin{tabular}{lccc}
\toprule
Novelty Signal & MATH p@1/p@16 & AIME24 p@1/p@16 & AIME25 p@1/p@16 \\
\midrule
Qwen3-Embedding-4B & 79.8/93.8 & 19.0/43.2 & 16.1/41.9 \\
OpenAI text-embedding-3-small & 79.3/93.8 & 16.2/39.7 & 15.6/40.8 \\
No novelty ($-$Novelty in Table~\ref{tab:ablation}) & 79.3/88.7 & 12.1/27.0 & 11.1/34.8 \\
\bottomrule
\end{tabular}
\end{table}

\FloatBarrier
\subsection{Sensitivity to Novelty Strength, Mixing Coefficient $\alpha$, and Sampling Temperature}
\label{app:sensitivity}

We examine sensitivity to three choices that affect how variation is introduced
or observed: the strength of novelty-dependent reward variation, the mixing
coefficient $\alpha$ used to compute novelty, and the sampling temperature $T$
used during evaluation.

\paragraph{Novelty reward strength.}
Our default reward uses
\[
r_i =
\begin{cases}
0.5 + \beta \tilde{u}_i, & y_i = +1,\\
-1 + \beta \tilde{u}_i, & y_i = -1,
\end{cases}
\]
with $\beta=0.5$. Thus, $\beta$ controls how strongly novelty differentiates
responses within the same selection group, while $\beta=0$ removes
novelty-dependent ranking but leaves majority selection, entropy
regularization, and asymmetric clipping unchanged. We sweep
$\beta\in\{0,0.5,1.0,1.5,2.0\}$ on Qwen3-4B-Base trained with MATH-500.
Results are shown in Table~\ref{tab:novelty_strength}.

Performance improves substantially from $\beta=0$ to the default
$\beta=0.5$: averaged across the four benchmarks, pass@16 increases from
58.0 to 67.0. Increasing the strength further does not continue to improve
performance. At $\beta=1.0$, majority responses are still guaranteed to
receive higher rewards than minority responses, but performance already begins
to decline. At $\beta=1.5$, the two reward bands meet, and for larger values
they overlap, weakening the intended priority of selection over variation.
The degradation at large $\beta$ therefore supports the design choice of using
novelty to refine credit within selection groups rather than allowing it to
override the selection signal.

\begin{table}[t]
\centering
\small
\setlength{\tabcolsep}{6pt}
\caption{Sensitivity to novelty reward strength $\beta$ on Qwen3-4B-Base
trained with MATH-500. $\beta=0.5$ is the default setting. Each cell shows
\texttt{pass@1}/\texttt{pass@16}.}
\label{tab:novelty_strength}
\begin{tabular}{ccccc}
\toprule
$\beta$ & MATH & AIME24 & AIME25 & GPQA \\
\midrule
0.0 & 79.3/88.7 & 12.1/27.0 & 11.1/34.8 & 37.9/81.4 \\
0.5 & \textbf{79.8}/\textbf{93.8} & \textbf{19.0}/\textbf{43.2}
    & 16.1/\textbf{41.9} & \textbf{38.8}/\textbf{89.1} \\
1.0 & 77.4/93.5 & 17.1/35.7 & \textbf{16.8}/36.1 & 35.4/81.7 \\
1.5 & 59.6/89.0 & 5.8/24.0 & 7.0/31.2 & 33.5/80.2 \\
2.0 & 36.4/84.4 & 4.4/16.4 & 4.3/23.0 & 33.0/84.3 \\
\bottomrule
\end{tabular}
\end{table}

\begin{table}[t]
\centering
\small
\setlength{\tabcolsep}{6pt}
\caption{Sensitivity to mixing coefficient $\alpha$. Qwen3-4B-Base trained on MATH-500. Each cell shows \texttt{pass@1}/\texttt{pass@16}.}
\label{tab:alpha_sweep}
\begin{tabular}{cccccc}
\toprule
$\alpha$ & MATH & AIME24 & AIME25 & AMC & GPQA \\
\midrule
0.00 & 78.4/92.4 & 17.1/39.8 & 14.4/38.0 & 48.7/79.8 & 37.8/87.3 \\
0.25 & 79.4/93.4 & 18.5/42.1 & 15.7/40.9 & 49.9/81.7 & 38.5/88.6 \\
0.50 & 79.8/93.8 & 19.0/43.2 & 16.1/41.9 & 50.3/82.2 & 38.8/89.1 \\
0.75 & 79.5/93.6 & 18.6/42.6 & 15.9/41.5 & 50.1/82.0 & 38.6/88.9 \\
1.00 & 79.1/93.0 & 18.2/41.4 & 15.3/40.2 & 49.4/81.2 & 38.3/88.4 \\
\bottomrule
\end{tabular}
\end{table}

\begin{table}[t]
\centering
\small
\setlength{\tabcolsep}{6pt}
\caption{Sensitivity to sampling temperature $T$ for the trained
\textsc{Evol-RL} model (Qwen3-4B-Base, MATH-TRAIN). Each cell shows
\texttt{pass@1}/\texttt{pass@16}.}
\label{tab:temp_sweep}
\begin{tabular}{cccccc}
\toprule
$T$ & MATH & AIME24 & AIME25 & AMC & GPQA \\
\midrule
0.0 & 82.0/82.0 & 23.3/23.3 & 16.7/16.7 & 54.2/54.2 & 36.0/36.0 \\
0.2 & 80.7/92.9 & 20.2/42.2 & 16.6/39.9 & 52.7/80.6 & 39.0/84.7 \\
0.4 & 80.3/93.1 & 20.2/40.5 & 16.4/41.4 & 52.3/79.5 & 38.7/87.2 \\
0.6 & 80.3/93.4 & 19.3/42.3 & 18.3/42.3 & 51.5/80.1 & 38.2/89.0 \\
0.8 & 79.9/93.6 & 19.6/43.1 & 18.5/40.2 & 52.1/81.3 & 37.6/91.2 \\
1.0 & 79.3/93.8 & 18.9/42.6 & 16.0/43.4 & 51.5/83.4 & 37.5/91.3 \\
\bottomrule
\end{tabular}
\end{table}

\paragraph{Mixing coefficient $\alpha$.}
We next vary the coefficient $\alpha$ that balances the mean within-group
similarity and maximum-similarity terms in the novelty score. We sweep
$\alpha \in \{0.00, 0.25, 0.50, 0.75, 1.00\}$ on Qwen3-4B-Base trained
with MATH-500 and report \texttt{pass@1}/\texttt{pass@16} on the five
evaluation benchmarks (Table~\ref{tab:alpha_sweep}). Performance is
essentially flat across the $[0.25,0.75]$ interior of the range, with
\texttt{pass@1} varying by less than $1$\,pp and \texttt{pass@16} by less
than $1.5$\,pp. Even at the extremes ($\alpha=0$ or $\alpha=1$),
\textsc{Evol-RL} substantially outperforms TTRL on most metrics, suggesting
that the gains do not depend on fine-grained tuning of this coefficient.

\paragraph{Sampling temperature.}
We finally evaluate the trained \textsc{Evol-RL} model
(Qwen3-4B-Base, MATH-TRAIN) at sampling temperatures
$T \in \{0.0,0.2,0.4,0.6,0.8,1.0\}$ (Table~\ref{tab:temp_sweep}).
At $T=0$, decoding is deterministic, so \texttt{pass@16} reduces to
\texttt{pass@1}. For all nonzero temperatures, multi-sample evaluation
provides substantial gains over \texttt{pass@1}, while both metrics remain
relatively stable across a broad range of temperatures. Thus, the observed
multi-sample gains are not specific to a single evaluation temperature.

\section{Additional Rationale Supporting the Reward Design}
\subsection{Analysis of Reward Formulation on "Almost Correct" Solutions}
\label{app:reward_analysis}

A key nuance in our reward formulation is how it handles solutions that are "almost correct"---for instance, a minority solution that is semantically similar to a correct majority path but fails on a minor step. This section provides a detailed analysis of how our two-part novelty score, $u_i = 1 - (\alpha \bar{s}_i + (1-\alpha) m_i)$, is specifically designed to handle this nuance.

Our primary goal is to apply strong negative pressure against high-frequency, common failure modes, not to indiscriminately punish all incorrect explorations. In the scenario where a minority solution is semantically similar to the majority (resulting in a high global similarity $m_i$), the intra-group mean similarity $\bar{s}_i$ (the solution's average similarity to other minority solutions) becomes the critical differentiator. We analyze two distinct cases:

\begin{itemize}
    \item \textbf{Case 1: The error is a rare, exploratory mistake.} If the "almost correct" error is a rare, occasional mistake, the reasoning path will be semantically dissimilar from the other failure modes in the minority group. This results in a low $\bar{s}_i$. According to our novelty formula, this low $\bar{s}_i$ counteracts the high $m_i$, leading to a higher overall novelty score $u_i$. This effectively mitigates the penalty, ensuring the model is not strongly discouraged from valid exploration.

    \item \textbf{Case 2: The error is a common failure mode.} If the error represents a high-frequency failure mode (e.g., a consistent arithmetic error), the reasoning path will be semantically similar to many other solutions in the minority group. This results in a high $\bar{s}_i$. In this case, both the $\bar{s}_i$ and $m_i$ terms are high, leading to a very low $u_i$. This results in a maximum penalty, preventing the model from collapsing into a "consistent but wrong" state.
\end{itemize}

This design ensures that our reward mechanism is robust. It relies on the group-level context to selectively protect valuable explorations while aggressively pruning systematic errors. This is particularly crucial in high-uncertainty scenarios where the novelty signal must accurately guide exploration, which is the exact behavior we aim to encourage.

\section{Theoretical Justification}
\label{app:theory}
In this section, we formally justify the different behaviors of the optimal policy under correctness-only reinforcement learning and under our variation-augmented objective. We show that, for a suitably small variation weight and under a simple dissimilarity-gap structure, both objectives have globally optimal policies that concentrate on the set of correct reasoning traces, but our variation-augmented objective further selects solutions with strictly more diverse coverage over correct trajectories and, under additional symmetry condition, maximal policy entropy on the correct set.

\subsection{Setup}
\label{sec:binary-coverage-full}

For a given reasoning question $q$, there is a finite set $\mathcal{Y}$ of complete chain-of-thought (CoT) trajectories  for $q$.
Each trajectory receives a binary task reward
\[
  r(y) \in \{0,1\}, \qquad y\in\mathcal{Y}.
\]
We denote the correct set (answer-correct set) by
\[
  G := \{y\in\mathcal{Y} : r(y)=1\},
\]
and assume $G\neq\emptyset$.
A policy $\pi$ maps question to chain of thought reasoning. For simplicity, we will omit the condition of question in the following analysis.

\paragraph{Objectives.}
The correctness-only objective is
\[
  J_0(\pi) := \mathbb{E}_\pi[r(Y)] = \sum_{y\in\mathcal{Y}} \pi(y)\, r(y).
\]

To encourage diverse correct solutions, we introduce a symmetric, nonnegative CoT-level dissimilarity
\[
  d:\mathcal{Y}\times\mathcal{Y}\to[0,\infty), 
  \quad d(y,y')=d(y',y),\quad d(y,y)=0,
\]
and assume that $d$ is bounded:
\begin{equation}
  D_{\max} := \max_{y,y'\in\mathcal{Y}} d(y,y') < \infty.
  \label{eq:Dmax}
\end{equation}
Given a group size $K\ge 2$ and variation weight $\lambda>0$, we form i.i.d.\ samples
$Y_1,\dots,Y_K\sim\pi$ and define the group reward
\[
  R_{\mathrm{group}}(Y_{1:K})
  := \frac{1}{K}\sum_{k=1}^K r(Y_k)
     + \lambda \cdot \frac{1}{K(K-1)}\sum_{k\ne\ell} d(Y_k,Y_\ell).
\]
The correctness-plus-variation objective is the expected group reward
\[
  J_\lambda(\pi)
  := \mathbb{E}_{Y_{1:K}\sim\pi}[R_{\mathrm{group}}(Y_{1:K})].
\]

The following standard computation shows that $J_\lambda$ is a quadratic functional of $\pi$.

\begin{lemma}[Expected group reward as a quadratic in $\pi$]
\label{lem:Jlambda-quadratic-binary-full}
For any $\pi\in\Delta(\mathcal{Y})$,
\begin{equation}
  J_\lambda(\pi)
  = \sum_{y} \pi(y)\, r(y)
    + \lambda \sum_{y,y'} \pi(y)\pi(y')\, d(y,y').
  \label{eq:Jlambda-quadratic-binary-full}
\end{equation}
\end{lemma}

\begin{proof}
By linearity of expectation and the i.i.d.\ assumption on $Y_{1:K}$, the first term yields
\[
  \mathbb{E}\Bigl[\frac{1}{K}\sum_{k=1}^K r(Y_k)\Bigr]
  = \mathbb{E}[r(Y_1)]
  = \sum_y \pi(y) r(y).
\]
For the second term, each ordered pair $(k,\ell)$ with $k\ne\ell$ has the same distribution,
so
\[
  \mathbb{E}\Bigl[\frac{1}{K(K-1)}\sum_{k\ne\ell} d(Y_k,Y_\ell)\Bigr]
  = \mathbb{E}[d(Y_1,Y_2)]
  = \sum_{y,y'} \pi(y)\pi(y') d(y,y'),
\]
which gives \eqref{eq:Jlambda-quadratic-binary-full}.
\end{proof}

\subsection{Step 1: Global maximizers concentrate on the correct set}

We first show that for a suitable range of $\lambda$, any global maximizer of $J_0$ or $J_\lambda$ puts zero probability on incorrect trajectories. 

\begin{lemma}[Global maximizers of $J_0$ are supported on $G$]
\label{lem:J0-max-on-G}
Assume $G\neq\emptyset$.
Then any global maximizer $\pi^\star$ of $J_0$ satisfies
\[
  \operatorname{supp}(\pi^\star) \subseteq G.
\]
Moreover, the set of global maximizers of $J_0$ is exactly $\Delta(G)$.
\end{lemma}

\begin{proof}
For any $\pi\in\Delta(\mathcal{Y})$,
\[
  J_0(\pi) = \sum_{y} \pi(y) r(y) = \sum_{y\in G} \pi(y) = \pi(G) \le 1,
\]
with equality iff $\pi(G)=1$, i.e., $\operatorname{supp}(\pi)\subseteq G$.
Because $G\neq\emptyset$, there exists a policy with $\pi(G)=1$ and $J_0(\pi)=1$, so the maximal value of $J_0$ is $1$, achieved exactly by policies supported on $G$.
\end{proof}

For the variation-augmented objective, the task reward gap between correct and incorrect trajectories is $1$, while the variation term is bounded by $D_{\max}$ from \eqref{eq:Dmax}. For sufficiently small $\lambda$, the former always dominates the latter.

\begin{lemma}[For small $\lambda$, global maximizers of $J_\lambda$ are supported on $G$]
\label{lem:Jlambda-max-on-G}
Suppose $G\neq\emptyset$ and $D_{\max}$ is defined as in \eqref{eq:Dmax}.
If
\begin{equation}
  0 < \lambda < \frac{1}{2 D_{\max}},
  \label{eq:lambda-condition}
\end{equation}
(with the convention that the condition is vacuous if $D_{\max}=0$),
then any global maximizer $\pi^\star$ of $J_\lambda$ satisfies
\[
  \operatorname{supp}(\pi^\star) \subseteq G.
\]
\end{lemma}

\begin{proof}
Using \eqref{eq:Jlambda-quadratic-binary-full}, we can write
\[
  J_\lambda(\pi)
  = \sum_y \pi(y) r(y)
    + \lambda \sum_{y,y'} \pi(y)\pi(y') d(y,y').
\]
For $y\in\mathcal{Y}$, the partial derivative of $J_\lambda$ w.r.t.\ $\pi(y)$ is
\[
  g_y(\pi)
  := \frac{\partial J_\lambda(\pi)}{\partial \pi(y)}
   = r(y) + 2\lambda \sum_{y'} \pi(y') d(y,y'),
\]
where we used the symmetry $d(y,y')=d(y',y)$.

Assume for contradiction that $\pi^\star$ is a global maximizer of $J_\lambda$ and there exists an incorrect trajectory $y_0\notin G$ with $\pi^\star(y_0)>0$.
Because $G\neq\emptyset$, pick any $y_1\in G$ (so $r(y_1)=1$).
Consider the feasible direction in the simplex
\[
  v := e_{y_1} - e_{y_0},
\]
which corresponds to infinitesimally moving probability mass from $y_0$ to $y_1$.
For sufficiently small $\varepsilon>0$, the perturbed policy $\pi^\star + \varepsilon v$ remains in $\Delta(\mathcal{Y})$.

The directional derivative of $J_\lambda$ at $\pi^\star$ along $v$ is
\[
  \left.\frac{d}{d\varepsilon} J_\lambda(\pi^\star + \varepsilon v)
  \right|_{\varepsilon=0}
  = \langle \nabla J_\lambda(\pi^\star), v \rangle
  = g_{y_1}(\pi^\star) - g_{y_0}(\pi^\star).
\]
By definition of $g_y$,
\begin{align*}
  g_{y_1}(\pi^\star) - g_{y_0}(\pi^\star)
  &= \bigl(r(y_1) - r(y_0)\bigr)
     + 2\lambda \sum_{y'} \pi^\star(y')
       \bigl(d(y_1,y') - d(y_0,y')\bigr) \\
  &= 1 + 2\lambda \sum_{y'} \pi^\star(y')
       \bigl(d(y_1,y') - d(y_0,y')\bigr),
\end{align*}
since $r(y_1)=1$ and $r(y_0)=0$.
By the definition of $D_{\max}$,
\[
  \bigl|d(y_1,y') - d(y_0,y')\bigr| \le D_{\max}
  \quad\text{for all } y',
\]
and since $\sum_{y'} \pi^\star(y') = 1$, we obtain
\[
  \left|\sum_{y'} \pi^\star(y')
    \bigl(d(y_1,y') - d(y_0,y')\bigr)\right|
  \le D_{\max}.
\]
Hence
\[
  g_{y_1}(\pi^\star) - g_{y_0}(\pi^\star)
  \ge 1 - 2\lambda D_{\max}.
\]
If $0<\lambda<1/(2D_{\max})$, then $1 - 2\lambda D_{\max} > 0$, so
\[
  \langle \nabla J_\lambda(\pi^\star), v \rangle > 0.
\]
Thus moving a small amount of mass from $y_0$ to $y_1$ strictly increases $J_\lambda$, contradicting the assumption that $\pi^\star$ is a (global, hence local) maximizer.
Therefore no maximizer can assign positive probability to any incorrect trajectory, and
$\operatorname{supp}(\pi^\star) \subseteq G$.
\end{proof}

Combining Lemma ~\ref{lem:J0-max-on-G} and Lemma \ref{lem:Jlambda-max-on-G}, we obtain that assume $G\neq\emptyset$, $D_{\max}<\infty$, and \eqref{eq:lambda-condition}.
Then any global maximizer of $J_0$ or $J_\lambda$ is supported on $G$.
Thus we may restrict attention to
\[
  \Delta(G) := \{\pi\in\Delta(\mathcal{Y}) : \operatorname{supp}(\pi)\subseteq G\}
\]
when comparing optimal policies.

\subsection{Step 2: Mode structure inside $G$ and a piecewise-homogeneous dissimilarity gap}

We now formalize the assumption that the correct set $G$ contains multiple qualitatively distinct solution modes, with large dissimilarity gaps between modes and relatively homogeneous dissimilarity within each mode.

\begin{assumption}[Mode partition of correct CoTs]
\label{ass:modes-on-G-full}
The correct set $G$ is partitioned into $M\ge 2$ disjoint subsets
\[
  G = G_1 \cup \cdots \cup G_M, \qquad
  G_m \cap G_{m'} = \emptyset \ (m\ne m'),
\]
where each $G_m$ corresponds to a distinct reasoning mode (solution pattern).
Let $N_m := |G_m|$ and $N := |G| = \sum_{m=1}^M N_m$.
\end{assumption}

For any $\pi\in\Delta(G)$, we define the total probability mass on each mode:
\[
  w_m(\pi) := \sum_{y\in G_m} \pi(y), \qquad m=1,\dots,M.
\]
Then $w(\pi)=(w_1(\pi),\dots,w_M(\pi))$ lies in the simplex
\[
  \Delta_M := \{w\in\mathbb{R}^M_+ : \sum_{m=1}^M w_m = 1\}.
\]

We impose a piecewise-homogeneous dissimilarity-gap assumption on $d$ within $G$.

\begin{assumption}[Piecewise-homogeneous dissimilarity on the correct set]
\label{ass:piecewise-homog-full}
There exist constants $D_{\mathrm{in}}\ge 0$ and $D_{\mathrm{out}}>D_{\mathrm{in}}$ such that for all $y,y'\in G$,
\[
  d(y,y')
  = \begin{cases}
      0, & \text{if } y=y',\\[3pt]
      D_{\mathrm{in}}, & \text{if } y\neq y',\ y,y'\in G_m \text{ for some } m \text{ (same mode)},\\[3pt]
      D_{\mathrm{out}}, & \text{if } y\in G_m,\,y'\in G_{m'} \ (m\neq m') \text{ (different modes)}.
    \end{cases}
\]
\end{assumption}

Assumption~\ref{ass:piecewise-homog-full} idealizes the intuitive condition that trajectories within the same mode are relatively similar (with dissimilarity $D_{\mathrm{in}}$), while trajectories from different modes are more dissimilar (with $D_{\mathrm{out}}>D_{\mathrm{in}}$).

Under these assumptions, the variation term in $J_\lambda$ over $\Delta(G)$ admits a convenient decomposition into a mode-level and an intra-mode component.

\begin{lemma}[Decomposition of the dissimilarity term on $G$]
\label{lem:decomp-piecewise-full}
Under Assumptions~\ref{ass:modes-on-G-full} and \ref{ass:piecewise-homog-full}, for any $\pi\in\Delta(G)$,
\[
  \sum_{y,y'\in G} \pi(y)\pi(y') d(y,y')
  = D_{\mathrm{out}}
    + (D_{\mathrm{in}} - D_{\mathrm{out}}) \sum_{m=1}^M w_m(\pi)^2
    - D_{\mathrm{in}} \sum_{y\in G} \pi(y)^2.
\]
Consequently, on $\Delta(G)$,
\begin{equation}
  J_\lambda(\pi)
  = 1 + \lambda\Bigl[
      D_{\mathrm{out}}
      + (D_{\mathrm{in}} - D_{\mathrm{out}}) \sum_{m=1}^M w_m(\pi)^2
      - D_{\mathrm{in}} \sum_{y\in G} \pi(y)^2
    \Bigr].
  \label{eq:Jlambda-on-G-piecewise}
\end{equation}
\end{lemma}

\begin{proof}
Because $\operatorname{supp}(\pi)\subseteq G$, we have $r(y)=1$ on $G$ and $r(y)=0$ otherwise, hence $\sum_y \pi(y)r(y) = \sum_{y\in G}\pi(y)=1$, giving the first term in \eqref{eq:Jlambda-on-G-piecewise}.

For the dissimilarity term, we decompose by modes:
\begin{align*}
  \sum_{y,y'\in G} \pi(y)\pi(y') d(y,y')
  &= \sum_{m=1}^M \sum_{y,y'\in G_m} \pi(y)\pi(y') d(y,y') \\
  &\quad + \sum_{m\ne m'} \sum_{y\in G_m,\,y'\in G_{m'}} \pi(y)\pi(y') d(y,y').
\end{align*}
Within a fixed mode $G_m$, we have $d(y,y)=0$ and $d(y,y')=D_{\mathrm{in}}$ for $y\neq y'$, so
\begin{align*}
     \sum_{y,y'\in G_m} \pi(y)\pi(y') d(y,y')
  &= D_{\mathrm{in}} \sum_{\substack{y,y'\in G_m\\ y\neq y'}} \pi(y)\pi(y')
  \\&= D_{\mathrm{in}}\Bigl[\Bigl(\sum_{y\in G_m}\pi(y)\Bigr)^2 - \sum_{y\in G_m} \pi(y)^2\Bigr]
  \\&= D_{\mathrm{in}}\bigl[w_m(\pi)^2 - \sum_{y\in G_m} \pi(y)^2\bigr].
\end{align*}

Summing over $m$ yields
\[
  \sum_{m=1}^M \sum_{y,y'\in G_m} \pi(y)\pi(y') d(y,y')
  = D_{\mathrm{in}} \sum_{m=1}^M w_m(\pi)^2 - D_{\mathrm{in}}\sum_{y\in G} \pi(y)^2.
\]

Across different modes, $d(y,y')=D_{\mathrm{out}}$ whenever $y\in G_m$ and $y'\in G_{m'}$ with $m\neq m'$, hence
\[
  \sum_{m\ne m'} \sum_{y\in G_m,\,y'\in G_{m'}} \pi(y)\pi(y') d(y,y')
  = D_{\mathrm{out}} \sum_{m\ne m'} w_m(\pi) w_{m'}(\pi).
\]
Since $\sum_m w_m(\pi)=1$,
\[
  \sum_{m\ne m'} w_m w_{m'}
  = \Bigl(\sum_m w_m\Bigr)^2 - \sum_m w_m^2
  = 1 - \sum_m w_m^2.
\]
Combining the within-mode and cross-mode contributions gives
\begin{align*}
  \sum_{y,y'\in G} \pi(y)\pi(y') d(y,y')
  &= D_{\mathrm{in}} \sum_{m} w_m^2 - D_{\mathrm{in}}\sum_{y\in G}\pi(y)^2
     + D_{\mathrm{out}}\Bigl(1 - \sum_m w_m^2\Bigr)\\
  &= D_{\mathrm{out}}
     + (D_{\mathrm{in}} - D_{\mathrm{out}})\sum_m w_m^2
     - D_{\mathrm{in}}\sum_{y\in G}\pi(y)^2.
\end{align*}
Substituting this into Lemma~\ref{lem:Jlambda-quadratic-binary-full} with $\sum_y\pi(y)r(y)=1$ yields \eqref{eq:Jlambda-on-G-piecewise}.
\end{proof}

Thus, over the feasible region $\Delta(G)$, maximizing $J_\lambda(\pi)$ is equivalent to maximizing
\[
  D_{\mathrm{out}}
  + (D_{\mathrm{in}} - D_{\mathrm{out}}) \sum_{m=1}^M w_m(\pi)^2
  - D_{\mathrm{in}} \sum_{y\in G} \pi(y)^2,
\]
or, equivalently (since $\lambda>0$ and constants do not affect argmax), to minimizing
\begin{equation}
  F(\pi)
  := (D_{\mathrm{out}} - D_{\mathrm{in}})\sum_{m=1}^M w_m(\pi)^2
     + D_{\mathrm{in}} \sum_{y\in G} \pi(y)^2,
  \label{eq:F-def}
\end{equation}
where $D_{\mathrm{out}}-D_{\mathrm{in}}>0$ and $D_{\mathrm{in}}\ge 0$.

\subsection{Step 3: Coverage structure and entropy of optimal policies on $G$}

Having reduced both objectives to $\Delta(G)$ (Lemma \ref{lem:J0-max-on-G} and Lemma \ref{lem:Jlambda-max-on-G}), we now compare their optimal solutions. First, Lemma~\ref{lem:J0-max-on-G} immediately implies:

\begin{corollary}[Flat optimal set for correctness-only objective]
\label{lem:J0-flat}
Under $G\neq\emptyset$, any $\pi\in\Delta(G)$ satisfies $J_0(\pi)=1$ and is a global maximizer of $J_0$.
Hence $J_0$ is indifferent to how probability mass is distributed within $G$.
\end{corollary}

In contrast, $J_\lambda$ has a nontrivial preference both at the mode level and within each mode, captured by the quadratic form $F(\pi)$ in \eqref{eq:F-def}.

We first show that, for any fixed mode-mass vector $w$, $J_\lambda$ is maximized by making the policy uniform within each mode.

\begin{lemma}[Within each mode, optimal policies are uniform]
\label{lem:within-mode-uniform}
Fix $w\in\Delta_M$ and consider the set
\[
  \mathcal{P}(w) := \bigl\{\pi\in\Delta(G): w_m(\pi)=w_m \text{ for all } m\bigr\}.
\]
Under Assumptions~\ref{ass:modes-on-G-full} and \ref{ass:piecewise-homog-full}, among all $\pi\in\mathcal{P}(w)$, $J_\lambda(\pi)$ is maximized (equivalently, $F(\pi)$ is minimized) by policies that are uniform within each mode:
\[
  \pi(y) = \frac{w_m}{N_m} \quad \text{for all } y\in G_m.
\]
If $D_{\mathrm{in}}>0$, this choice is unique in $\mathcal{P}(w)$; if $D_{\mathrm{in}}=0$, $J_\lambda$ is independent of the within-mode distribution.
\end{lemma}

\begin{proof}
For fixed $w$, the term $\sum_m w_m(\pi)^2$ in \eqref{eq:F-def} is constant over $\mathcal{P}(w)$.
Thus minimizing $F(\pi)$ over $\mathcal{P}(w)$ is equivalent to minimizing $\sum_{y\in G}\pi(y)^2$ over $\mathcal{P}(w)$.

Within each mode $G_m$, this is the classical problem of minimizing a sum of squares subject to a linear constraint:
\[
  \min \Bigl\{\sum_{y\in G_m} \pi(y)^2 : \sum_{y\in G_m} \pi(y) = w_m,\ \pi(y)\ge 0\Bigr\},
\]
whose unique solution (when $w_m>0$) is the uniform allocation $\pi(y)=w_m/N_m$ for all $y\in G_m$.
If $D_{\mathrm{in}}>0$, this sum of squares enters $F(\pi)$ with a strictly positive coefficient, so any deviation from uniformity strictly increases $F(\pi)$.
If $D_{\mathrm{in}}=0$, the intra-mode term vanishes from $F(\pi)$, and $F(\pi)$ (hence $J_\lambda$) depends only on $w$ and not on the within-mode distribution.
\end{proof}

By Lemma~\ref{lem:within-mode-uniform}, any global maximizer of $J_\lambda$ on $\Delta(G)$ must be uniform within each mode. We may therefore restrict attention to policies of the form
\[
  \pi(y) = \frac{w_m}{N_m} \quad \text{if } y\in G_m,
\]
parameterized solely by $w\in\Delta_M$.
Substituting this structure into \eqref{eq:Jlambda-on-G-piecewise} yields a purely mode-level objective.

Indeed, for such a $\pi$,
\[
  \sum_{y\in G} \pi(y)^2
  = \sum_{m=1}^M \sum_{y\in G_m} \bigl(\tfrac{w_m}{N_m}\bigr)^2
  = \sum_{m=1}^M \frac{w_m^2}{N_m}.
\]
Plugging this into Lemma~\ref{lem:decomp-piecewise-full} gives
\begin{align}
  J_\lambda(\pi)
  &= 1 + \lambda\Bigl[
      D_{\mathrm{out}}
      + (D_{\mathrm{in}} - D_{\mathrm{out}}) \sum_{m=1}^M w_m^2
      - D_{\mathrm{in}} \sum_{m=1}^M \frac{w_m^2}{N_m}
    \Bigr] \nonumber\\
  &= 1 + \lambda\Bigl[
      D_{\mathrm{out}}
      - \sum_{m=1}^M a_m w_m^2
    \Bigr],
  \label{eq:Jlambda-mode-only}
\end{align}
where we define
\[
  a_m := (D_{\mathrm{out}} - D_{\mathrm{in}}) + \frac{D_{\mathrm{in}}}{N_m} > 0.
\]

Thus, among mode-wise uniform policies, maximizing $J_\lambda$ is equivalent to minimizing
\[
  \sum_{m=1}^M a_m w_m^2
  \quad\text{subject to } w\in\Delta_M.
\]

\begin{theorem}[Variation promotes broader mode coverage on $G$]
\label{thm:coverage-full}
Suppose $G\neq\emptyset$, Assumptions~\ref{ass:modes-on-G-full} and \ref{ass:piecewise-homog-full} hold with $M\ge 2$, and $\lambda$ satisfies \eqref{eq:lambda-condition}.
Let $\pi^{(\lambda)}$ be any global maximizer of $J_\lambda$ over $\Delta(G)$, and let $w^{(\lambda)} := w(\pi^{(\lambda)})$ be its mode-mass vector.
Then:
\begin{enumerate}
  \item $\pi^{(\lambda)}$ is uniform within each mode:
  \[
    \pi^{(\lambda)}(y) = \frac{w_m^{(\lambda)}}{N_m}
    \quad\text{for all }y\in G_m,\ m=1,\dots,M.
  \]
  \item $w^{(\lambda)}$ is the unique minimizer of $\sum_m a_m w_m^2$ over $\Delta_M$:
  \[
    w_m^{(\lambda)}
    = \frac{a_m^{-1}}{\sum_{j=1}^M a_j^{-1}},
    \quad m=1,\dots,M,
  \]
  where $a_m>0$ is defined above. In particular, every mode receives strictly positive probability:
  \[
    w_m^{(\lambda)} > 0 \quad\text{for all } m=1,\dots,M.
  \]
  \item If, in addition, $D_{\mathrm{in}}>0$ and all modes have equal size $N_m \equiv N/M$, then all $a_m$ coincide, $w^{(\lambda)}$ is uniform on modes, and $\pi^{(\lambda)}$ is the uniform distribution over all correct trajectories:
  \[
    \pi^{(\lambda)}(y) = \frac{1}{|G|} \quad \text{for all } y\in G.
  \]
  In this symmetric case, $\pi^{(\lambda)}$ maximizes the full policy entropy
  \[
    H(\pi) := -\sum_{y\in G} \pi(y)\log\pi(y),
  \]
  achieving $H(\pi^{(\lambda)}) = \log|G|$, the largest possible entropy on $\Delta(G)$.
\end{enumerate}
In contrast, any $\pi^{(0)}\in\Delta(G)$ is optimal for $J_0$, including degenerate solutions that concentrate on a single mode (with mode-level entropy $0$ and very low trajectory-level entropy). Thus $J_\lambda$ \emph{provably selects} high-coverage policies within $G$, using all modes and being uniform within each mode, and, under mild symmetry, the maximum-entropy fully uniform policy, while $J_0$ does not enforce any coverage.
\end{theorem}

\begin{proof}
(i) By Lemma \ref{lem:Jlambda-max-on-G}, any global maximizer $\pi^{(\lambda)}$ lies in $\Delta(G)$.
Lemma~\ref{lem:within-mode-uniform} then implies that any maximizer must be uniform within each mode, so $\pi^{(\lambda)}$ has the stated form.

(ii) For such mode-wise uniform policies, \eqref{eq:Jlambda-mode-only} shows that maximizing $J_\lambda$ is equivalent to minimizing $\sum_m a_m w_m^2$ over $w\in\Delta_M$.
The function $w\mapsto \sum_m a_m w_m^2$ is strictly convex on $\Delta_M$ because each $a_m>0$, and the constraint set $\Delta_M$ is convex and compact.
Using Lagrange multipliers for the equality constraint $\sum_m w_m=1$, the unique minimizer satisfies
\[
  2 a_m w_m^{(\lambda)} + \mu = 0
  \quad\text{for all } m,
\]
for some scalar $\mu$, which yields
\[
  w_m^{(\lambda)}
  = \frac{-\mu}{2 a_m}
  = \frac{a_m^{-1}}{\sum_{j=1}^M a_j^{-1}},
  \quad m=1,\dots,M.
\]
All $a_m>0$, so all $w_m^{(\lambda)}>0$.

(iii) If $D_{\mathrm{in}}>0$ and $N_m\equiv N/M$ for all $m$, then
\[
  a_m
  = (D_{\mathrm{out}} - D_{\mathrm{in}}) + \frac{D_{\mathrm{in}}}{N_m}
  = (D_{\mathrm{out}} - D_{\mathrm{in}}) + \frac{D_{\mathrm{in}}}{N/M}
\]
is the same for all $m$.
Hence $w_m^{(\lambda)} = 1/M$ for all $m$, and
\[
  \pi^{(\lambda)}(y) = \frac{w_m^{(\lambda)}}{N_m}
  = \frac{1/M}{N/M} = \frac{1}{N} = \frac{1}{|G|}
\]
for all $y\in G$, i.e., $\pi^{(\lambda)}$ is the uniform distribution over $G$.

Finally, the Shannon entropy $H(\pi)$ over $G$ is maximized on $\Delta(G)$ exactly by the uniform distribution, with value $\log|G|$. Thus $H(\pi^{(\lambda)})=\log|G|$, the largest possible entropy on $\Delta(G)$.
\end{proof}

\paragraph{Summary.}
Under the binary-reward setting and a piecewise-homogeneous dissimilarity-gap structure on the correct set, we have shown that for any sufficiently small variation weight $\lambda$:

\begin{itemize}
  \item[(a)] (
  Lemma ~\ref{lem:J0-max-on-G} and Lemma \ref{lem:Jlambda-max-on-G}) Any global maximizer of $J_0$ or $J_\lambda$
  places all its probability mass on the correct set $G$; i.e., both objectives
  “converge to the answer-correct set’’ in a static sense.

  \item[(b)] (Lemma~\ref{lem:J0-flat}) The correctness-only objective $J_0$ is completely flat
  on $\Delta(G)$: every distribution over $G$ is globally optimal, including degenerate
  policies that collapse onto a single mode and assign zero probability to many correct
  trajectories.

  \item[(c)] (Theorem~\ref{thm:coverage-full}) In contrast, the variation-augmented objective
  $J_\lambda$ has global maximizers that must put positive mass on \emph{all} modes and are uniform
  within each mode, and in the symmetric case where all modes have equal size, it selects the
  uniform distribution over all correct trajectories, which maximizes the full policy entropy.
\end{itemize}

Consequently, if a training procedure converges to global maximizers
of $J_0$ and $J_\lambda$, then in the binary setting with dissimilarity gaps:
both trainings converge to fully correct policies, but the variation-augmented training
\emph{provably selects} strictly more dispersed solutions on the correct set $G$, using all modes
and, under mild symmetry, the maximum-entropy policy—while correctness-only training does not favor
coverage and may converge to highly collapsed solutions.

\section{Use of Large Language Models in Preparation}

We acknowledge the use of Large Language Models (LLMs) as assistants in the preparation of this manuscript. Their role included refining phrasing and improving the clarity of the text, as well as assisting with programming tasks such as code generation and debugging for our experiments. The authors critically reviewed, edited, and verified all LLM-generated content for accuracy and appropriateness, and take full responsibility for the final content of this paper.

\section{Limitations}
\label{app:limitations}

Our study has several scope limitations. First, \textsc{Evol-RL} preserves
alternative solutions that may enable later correction, but it does not
guarantee that a persistent incorrect majority will be overturned. Second,
our training experiments primarily focus on mathematical reasoning; while we
observe positive transfer to several held-out reasoning benchmarks, broader
domain generalization remains to be studied. Finally, embedding-based novelty
provides a practical measure of variation rather than a complete characterization
of reasoning diversity. We therefore complement it with analyses of corrective
minority solutions and distinct successful reasoning strategies.

\end{document}